\documentclass[10pt,journal,compsoc]{IEEEtran}

\ifCLASSOPTIONcompsoc
  \usepackage[nocompress]{cite}
\else
  \usepackage{cite}

\fi

\ifCLASSINFOpdf
\else
\fi

\usepackage{amsmath}
\usepackage{acronym}
\usepackage{algorithmic}
\usepackage{booktabs}
\usepackage{multirow}
\usepackage{amssymb} 
\usepackage{amssymb} 
\usepackage{booktabs} 
\usepackage{array}    
\usepackage{pgfplots}
\pgfplotsset{compat=1.17} 

\newcommand{\cmark}{\checkmark}

\usepackage{genealogytree} 
\usepackage{tabularx}
\usepackage{xspace}
\usepackage{cite}
\usepackage{hyperref}
\usepackage{natbib}
\usepackage{graphicx}
\usepackage{float}
\usepackage{subfigure}
\usepackage{forest}
\usepackage{url}
\usepackage{todonotes}
\usepackage{xcolor}
\usepackage{comment}
\usepackage{tikz}
\usepackage{pgfplots}
\usepackage{pgfplotstable}
\usepackage{booktabs} 
\usepackage{adjustbox}
\usepackage{wrapfig}

\newcommand{\yly}[2]{{\color{orange}{[(YLY): ]}}}

\begin{document}

\title{A Survey on Data Augmentation in Large Model Era}

\author{Yue Zhou,
        Chenlu Guo,
        Xu Wang,
        Yi Chang,~\IEEEmembership{Senior Member,~IEEE,}
        Yuan Wu,~\IEEEmembership{Member,~IEEE}
\IEEEcompsocitemizethanks{\IEEEcompsocthanksitem Y. Zhou, C. Guo, X. Wang, Y. Wu and Y. Chang are with the School of Artificial Intelligence, Jilin University, Changchun, China. The first two authors contributed equally. 
\IEEEcompsocthanksitem Correspondence to: Yuan Wu (yuanwu@jlu.edu.cn).}
\thanks{Manuscript received April 19, 2005; revised August 26, 2015.}}

\markboth{Journal of \LaTeX\ Class Files,~Vol.~14, No.~8, August~2015}%
{Shell \MakeLowercase{\textit{et al.}}: Bare Advanced Demo of IEEEtran.cls for IEEE Computer Society Journals}
%


\IEEEtitleabstractindextext{%
\begin{abstract}

Large models, encompassing large language and diffusion models, have shown exceptional promise in approximating human-level intelligence, garnering significant interest from both academic and industrial spheres. However, the training of these large models necessitates vast quantities of high-quality data, and with continuous updates to these models, the existing reservoir of high-quality data may soon be depleted. This challenge has catalyzed a surge in research focused on data augmentation methods. Leveraging large models, these data augmentation techniques have outperformed traditional approaches. This paper offers an exhaustive review of large model-driven data augmentation methods, adopting a comprehensive perspective. We begin by establishing a classification of relevant studies into three main categories: image augmentation, text augmentation, and paired data augmentation. Following this, we delve into various data post-processing techniques pertinent to large model-based data augmentation. Our discussion then expands to encompass the array of applications for these data augmentation methods within natural language processing, computer vision, and audio signal processing. We proceed to evaluate the successes and limitations of large model-based data augmentation across different scenarios. Concluding our review, we highlight prospective challenges and avenues for future exploration in the field of data augmentation. Our objective is to furnish researchers with critical insights, ultimately contributing to generating sufficient and diverse data to train more sophisticated large models. We consistently maintain the related open-source materials at: \url{https://github.com/MLGroup-JLU/LLM-data-aug-survey}.

\end{abstract}

}

\maketitle

\IEEEdisplaynontitleabstractindextext

\IEEEpeerreviewmaketitle


\ifCLASSOPTIONcompsoc
\IEEEraisesectionheading{\section{Introduction}\label{sec:introduction}}
\else
\section{Introduction}
\label{sec:introduction}
\fi

\IEEEPARstart{D}{ata} augmentation, a pivotal strategy in machine learning, addresses the challenge of training models with limited labeled data for diverse tasks. It involves enhancing the sufficiency and diversity of training examples without explicitly collecting new data, thus playing a crucial role in improving model generalization \citep{shorten2019survey,feng2021survey}. The essence of data augmentation lies in generating new data by altering existing data points through various transformations. This prevents models from memorizing irrelevant data patterns, with the augmented data closely mirroring the distribution of real data \citep{wei2019eda,cubuk2019autoaugment}. Such techniques are directly applicable in supervised learning \citep{liu2021adaptive} and can be employed in semi-supervised learning for unlabeled data through consistency regularization \citep{zhang2021flexmatch}. Originally developed for computer vision (CV), data augmentation methods create artificial images through operations like cropping, rotating, and color adjustment \citep{takahashi2019data,krell2017rotational,kanwal2022devil}. In natural language processing (NLP), similar approaches involve random character insertion, word deletion, and synonym replacement \citep{shorten2019survey,liu2020survey}.

The significance of data augmentation has attracted substantial attention in both academic and industrial fields. As a vibrant research area, it addresses the growing need for large volumes of high-quality labeled data in machine learning, a demand often unmet in real-world scenarios. Despite significant advancements in data augmentation over the past decades, especially with deep learning techniques, these methods still struggle with capturing the complexities of real-world data \citep{feng2021survey}, generating scalable data \citep{yang2022image}, and defending against adversarial examples \citep{qiu2020fencebox}.

In response to these limitations, current research is exploring innovative techniques to enhance the efficacy and diversity of data augmentation methods. Among these, large models, including large language models \citep{zhao2023survey} and diffusion models \citep{yang2023diffusion}, show considerable promise. Large language models (LLMs), such as GPT-4 \citep{openai2023gpt4} and Llama2 \citep{touvron2023llama2}, have revolutionized NLP. Characterized by transformer architectures \citep{vaswani2017attention} and trained on extensive corpora, LLMs excel in understanding and generating human-like text, marking a significant advancement in machine learning capabilities \citep{zhao2023survey}. These models, with billions of parameters, can undertake diverse and complex tasks, including code generation \citep{zhang2023planning} and data augmentation \citep{dai2023chataug}, paving the way towards artificial general intelligence (AGI).

Diffusion models \citep{ho2020denoising,song2020score}, a new family of state-of-the-art generative models, have surpassed the long-standing dominance of generative adversarial networks (GANs) \citep{goodfellow2014explaining} in image synthesis within computer vision \citep{dhariwal2021diffusion,ho2020denoising}. Unlike prior models like variational auto-encoders (VAEs) \citep{kingma2013auto} and GANs, diffusion models iteratively add and reverse noise to generate high-quality synthetic images and have enabled text-to-image generation \citep{saharia2022photorealistic}, expanding the scope of data augmentation.

With the impressive capabilities of LLMs and diffusion models, there is a growing interest in using these models for data augmentation in both NLP and CV. This has led to the creation of more diverse and comprehensive datasets \citep{dai2023chataug,trabucco2023effective,sahu2022data,samuel2023can}. Over the past year, research on large model-based data augmentation has advanced significantly, posing a challenge for new researchers to keep pace with recent developments and discern major trends. A systematic summary of this rapidly evolving field is therefore crucial to provide a comprehensive understanding and inspire future research.

\begin{figure*}[t!]
	\centering
	\resizebox{\textwidth}{!}{
	\begin{forest}
  for tree={
  grow=east,
  reversed=true,
  anchor=base west,
  parent anchor=east,
  child anchor=west,
  base=left,
  edge path={ 
      \noexpand\path [draw, \forestoption{edge}] (!u.parent anchor) -- +(5pt,0) |- (.child anchor)\forestoption{edge label};
    },
  font=\small,
  rectangle,
  draw,
  rounded corners,align=left,
  minimum width=2.5em,
  inner xsep=4pt,
  inner ysep=1pt,
  },
  where level=1{fill=green!20}{}, 
  where level=2{font=\footnotesize,fill=cyan!15}{},
  where level=3{font=\footnotesize,fill=yellow!20}{},
  [Large Model-based \\Data Augmentation,fill=red!20 
        [Approach\\(Sec. \ref{sec-appro})
            [Image Augmentation
                [Prompt-driven
                    [Text Prompt-driven,fill=pink!30
                        [\citep{luo2023camdiff};\citep{couairon2022diffedit};\citep{nichol2021glide};\\ \citep{samuel2023all};\citep{tumanyan2023plug};\citep{hertz2022prompt};\\ \citep{zhang2023sine};\citep{bar2022text2live};\citep{kim2022diffusionclip};\\ \citep{gal2022stylegan};\citep{couairon2022diffedit};\\ \citep{dunlap2023diversify};\citep{trabucco2023effective};\citep{kawar2023imagic};\\ \citep{yin2023ttida};\citep{avrahami2022blended};\citep{wang2023dynamic};\\ \citep{brooks2023instructpix2pix};\citep{ge2023expressive}\cite{patashnik2023localizing};\\
                        \citep{koohpayegani2023genie};\citep{doubinsky2023semantic};\citep{feng2024instagen};\\
                        \citep{hemati2023cross};\citep{feng2024instagen};\citep{hemati2023cross}
                        ]
                    ]
                    [Visual Prompt-driven,fill=pink!30
                        [
                        \citep{sun2023imagebrush};\citep{yu2023diffusion};\citep{wu2023image};\citep{liu2023more} 
                        ]
                    ]
                    [Multimodal Prompt-driven,fill=pink!30
                        [\citep{nguyen2023visual};\citep{wang2023context};\citep{xie2023smartbrush};\\ \citep{huang2023reversion};\citep{du2023boosting};\citep{schnell2023generative};\\
                        \citep{zhang2023adding};\citep{li2023gligen};\citep{erfanian2024chameleon};\\
                        \citep{wang2024diffusion};\citep{erfanian2024chameleon};\citep{wang2024diffusion}
                        ]
                    ]                 
               ]
               [Subject-driven
                   [              \citep{gal2022image};\citep{li2023blip};\citep{xiao2023fastcomposer};\citep{chen2023subject};\\ \citep{wei2023elite};\citep{shi2023instantbooth};\citep{kumari2023multi};\citep{ma2023unified}
                   ]
               ]
            ]       
            [Text Augmentation
                [Label-based
                    [                  \citep{thakur2020augmented};\citep{kumar2020data};\citep{sahu2022data};\citep{yoo2021gpt3mix};\\ \citep{sharma2023augmenting};\citep{samuel2023can};\citep{latif2023can};\citep{chowdhury2023generative};\\ \citep{chen2023minprompt};\citep{kaddour2023text};\citep{wu2023improving};\citep{ye2024llm};\citep{ye2024llm}
                    ]
                ]
                [Generated Content-based
                    [  \citep{zheng2023augesc};\citep{dai2023chataug};\citep{yu2022coca};\citep{jo2022dagam};\\ \citep{oh2023data};\citep{tarjan2020deep};\citep{zhou2021flipda};\citep{guo2022genius};\\ \citep{bonifacio2022inpars};\citep{khatri2022skillbot};\citep{quteineh2020textual};\\ \citep{liu2022wanli};\citep{lu2023epa};\citep{meng2023tuning};\citep{meng2022generating};\\ \citep{saakyan2023iclef};\citep{ko2023natural};\citep{schlegel2023pulsar};\\ \citep{gao2022self};\citep{cai2023resolving};\citep{tang2023just};\citep{li2023dail};\\
                    \citep{ubani2023zeroshotdataaug};\citep{wang2024large};\citep{wang2024large}
                    ]
                ]
            ]
            [Paired Data Augmentation
                [       \citep{hao2023mixgen};\citep{bakhtiarnia2023promptmix};\citep{wu2023towards},fill=white!20
                ]
            ]    
        ]
        [Data Post Processing\\(Sec. \ref{sec-DPP})
            [Top-K Selection
                [    \citep{bonifacio2022inpars};\citep{meng2022generating};\citep{vu2021strata},fill=white!20
                ]
            ]
            [Model-based Approaches
                [              \citep{sachdeva2023catfood};\citep{samuel2023can};\citep{sharma2023augmenting};\citep{sahu2022data};\citep{zhou2021flipda},fill=white!20
                ]
            ]
            [Score-based Approaches
                [    \citep{shao2023diffuseexpand};\citep{wu2023image};\citep{chowdhury2023generative};\citep{zheng2023augesc};\citep{liu2022wanli},fill=white!20
                ]
            ]
            [Cluster-based Approaches
                [
                \citep{yu2023diffusion},fill=white!20
                ]
            ]
        ]
        [Application\\(Sec. \ref{sec-app})
            [Natural Language Processing
                [Text Classification
                    [                 \citep{dai2023chataug};\citep{jo2022dagam};\citep{kumar2020data};\\ \citep{yoo2021gpt3mix};\citep{gao2022self};\citep{cai2023resolving};\\ \citep{guo2022genius};\citep{meng2023tuning};\citep{saakyan2023iclef};\\
                    \citep{li2023dail}
                    ] 
                ] 
                [Question Answering
                    [           \citep{chen2023minprompt};\citep{chowdhury2023generative};\\ \citep{samuel2023can};\citep{sachdeva2023catfood}
                    ] 
                ]
                [Machine Translation
                    [
                    \citep{lu2023epa};\citep{oh2023data}
                    ] 
                ]
                [Natural Language Inference
                    [
                    \citep{lu2023epa};\citep{liu2022wanli}
                    ] 
                ]
                [Dialogue Summarising
                    [
                    \citep{lu2023epa};\citep{schlegel2023pulsar}
                    ] 
                ]
                [Others 
                    [   \citep{zheng2023augesc};\citep{thakur2020augmented};\citep{bonifacio2022inpars};\citep{lu2023epa}
                    ] 
                ]
            ]
            [Computer Vision
                [Image Classification
                    [                    \citep{samuel2023all};\citep{dunlap2023diversify};\citep{trabucco2023effective};\citep{yin2023ttida};\\ 
                    \citep{zang2023boosting};\citep{koohpayegani2023genie}
                    ] 
                ] 
                [Semantic Segmentation
                    [  \citep{zhang2023emit};\citep{du2023boosting};\citep{yu2023diffusion};\citep{wu2023image};\\
                    \citep{schnell2023generative}
                    ] 
                ]
                [Object Detection
                    [
                    \citep{voetman2023big};\citep{lu2023wovogen}
                    ] 
                ]
            ]
            [Audio Signal Processing
                [        \citep{wu2023improving};\citep{sharma2023augmenting};\citep{latif2023can};\citep{tarjan2020deep};\citep{kim2023adversarial},fill=white!20
                ]
            ]
        ]
        [Summary\\(Sec. \ref{sec-sum})
            [The successes and failures of large model-based data augmentation
                [Achievements]
                [Limitations]
            ]
            [Protocols and benchmarks for evaluation
                [Using performance metrics of downstream tasks:\\ \citep{dai2023chataug};\citep{jo2022dagam};\citep{yoo2021gpt3mix}; \citep{chowdhury2023generative};\\ \citep{samuel2023can};\citep{sachdeva2023catfood};\citep{oh2023data,lu2023epa};\citep{oh2023data};\\ \citep{lu2023epa};\citep{schlegel2023pulsar};\citep{lu2023epa};\citep{thakur2020augmented};\\ \citep{dunlap2023diversify};\citep{trabucco2023effective};\citep{yin2023ttida};\citep{voetman2023big};\\ \citep{lu2023wovogen};\citep{zhang2023emit};\citep{du2023boosting};\citep{yu2023diffusion};\\ \citep{wu2023image};\citep{sharma2023augmenting};\citep{tarjan2020deep};\citep{wu2023improving},fill=white!20
                ]
                [Using certain quality metrics of generated data:\\ \citep{dai2023chataug};\citep{ubani2023zeroshotdataaug};\citep{heusel2017gans};\citep{radford2021learning};\\ \citep{caron2021emerging};\citep{zhang2023adding};\citep{kumari2023multi};\citep{shi2023instantbooth};\\ \citep{ge2023expressive};\citep{brooks2023instructpix2pix};\citep{tumanyan2023plug};\citep{ruiz2023dreambooth};\\ \citep{tumanyan2023plug};\citep{wei2023elite},fill=white!20
                ]
            ]
        ]
        [Grand Challenges\\(Sec. \ref{sec-challenge})
            [(1) Theoretical understanding; (2) The number of augmented data; (3)Multimodal data augmentation; \\
            (4) Language and vision foundation models; (5) Automatic data augmentation;\\ 
            (6) Robust and consistent data augmentation; (7) Trustworthy data augmentation; \\
            (8)The instruction following ability of large models
            (9) The evaluation of augmented data;\\(10) Beyond augmentation: Training large models using augmented data] 
        ]
    ]
\end{forest}
	}
	\caption{Structure of this paper.}
	\label{fig-structure}
\end{figure*}
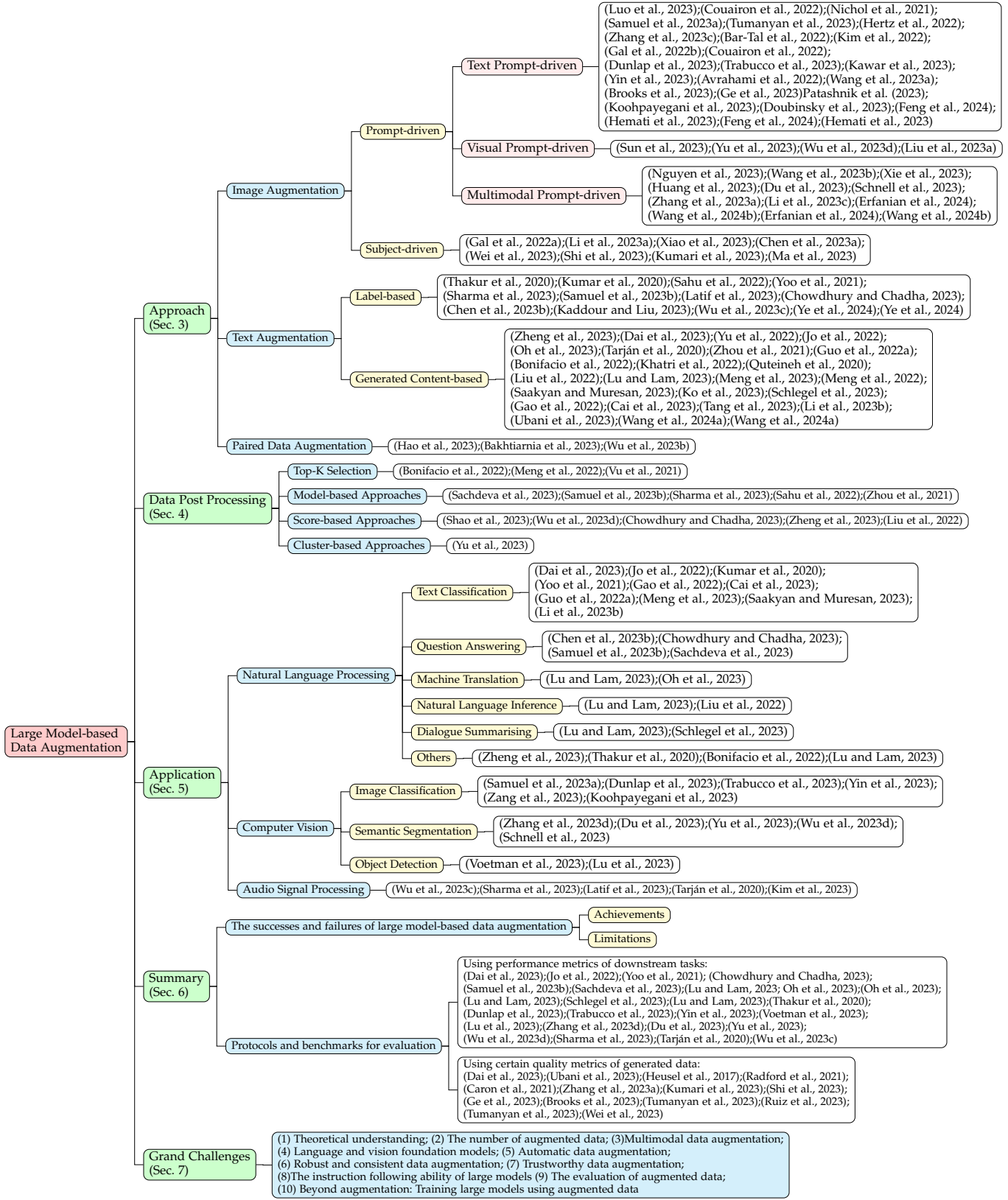

In this paper, we present an extensive survey of large model-based data augmentation approaches. As outlined in \figurename~\ref{fig-structure}, we structure our survey across three dimensions: approach, data post-processing, and application. The "approach" dimension includes image, text, and paired data augmentation methods; "data post-processing" covers methodologies like Top-$K$ selection, model-based, score-based, and cluster-based approaches; and "application" offers insights into the use of large model-based data augmentation in NLP, CV, and audio signal processing. This survey provides a systematic overview, establishes comprehensive taxonomies, identifies challenges, and discusses potential future directions in the field of large model-based data augmentation.

\begin{figure*}[htbp]
    \centering
    \includegraphics[width=\textwidth]{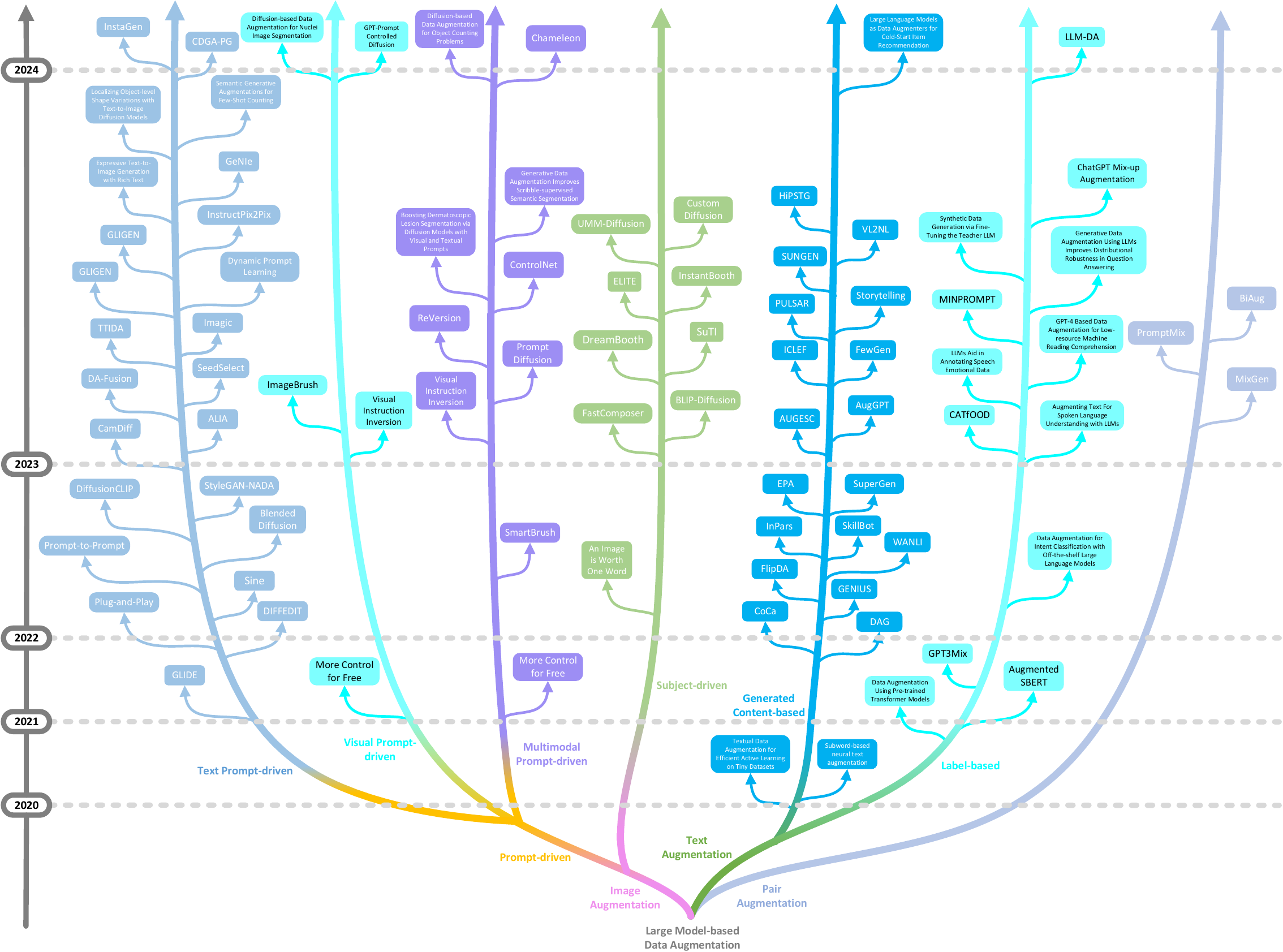}
    \caption{An evolutionary tree of large model-based data augmentation methods.}
    \label{fig:evolutionary-tree}
\end{figure*}

The contributions of this paper are as follows:

\begin{enumerate}
    \item We present a comprehensive overview of large model-based data augmentation methods, spanning three dimensions: approach, data post-processing, and application. Our analysis covers data augmentation techniques applicable to natural language processing (NLP), computer vision (CV), and audio signal processing.
    \item In terms of approach, we extensively summarize existing data augmentation methods that leverage advanced large language models (LLMs) and diffusion models. This provides an insightful perspective into the future trajectory of data augmentation research, we also show the evolutionary tree of large model-based data augmentation in \figurename~\ref{fig:evolutionary-tree}.
    \item Regarding data post-processing, we delineate Top-$K$ selection, model-based, score-based, and cluster-based approaches, aiming to explore how to refine augmented data. As for application, we examine the deployment of large model-based data augmentation methods across various tasks and obtain valuable conclusions on the success and failure cases of large model-based data augmentations.
    \item We summarize current protocols and benchmarks to evaluate large model-based data augmentation methods. We also delve into future challenges associated with the development of data augmentation methods that incorporate sophisticated large models.
\end{enumerate}

The structure of this paper is as follows: Section \ref{sec-back} introduces the foundational concepts of LLMs, diffusion models, and data augmentation methods. Section \ref{sec-appro} reviews current large model-based data augmentation methods, focusing on (1) Image data augmentation, (2) Text data augmentation, and (3) Paired data augmentation techniques. Section \ref{sec-DPP} examines existing data post-processing approaches that utilize large models. In Section \ref{sec-app}, we outline the application scenarios of the surveyed papers. Section \ref{sec-sum} synthesizes the key insights from this survey, while Section \ref{sec-challenge} explores significant future challenges in the field. The paper concludes with Section \ref{sec-conclusion}, summarizing our findings and contributions.

\section{Background}
\label{sec-back}
\subsection{Large Language Models}

The evolution of large language models (LLMs) can be traced to the early era of statistical language models \citep{rosenfeld2000two,liu2005statistical,gao2004introduction}, which estimated the probability distribution of linguistic elements like words, sentences, and documents, based on the Markov assumption. N-gram models \citep{brown1992class,o1994weighted,marino2006n}, the most prevalent among statistical language models, predict the likelihood of a word based on its preceding $n-1$ words. Despite their widespread use, these models struggled to capture long-range dependencies and intricate linguistic structures.

The advent of deep learning catalyzed the emergence of neural language models, utilizing neural networks \citep{bengio2000neural} to predict word sequence probabilities. Among these, Recurrent Neural Networks (RNNs) \citep{mikolov2010recurrent} emerged as a significant improvement over n-gram models. However, RNNs had limitations in handling long sequence data, as they could only capture information from nearby sequences and often lost memory of earlier sequences. This led to the development of Long Short Term Memory networks (LSTMs) \citep{sundermeyer2012lstm}, which addressed this limitation through a gating mechanism and cell states, enabling the capture of long-term dependencies. These models have shown significant improvements over traditional statistical models, but they still suffer from limitations in their ability to capture context and long-range dependencies.

The game-changing introduction of Transformer architecture with self-attention mechanisms marked a significant milestone \cite{vaswani2017attention}. Initially designed for machine translation, Transformers could capture both long-range dependencies and contextual information. This breakthrough spurred the development of pre-trained language models (PLMs) like BERT, GPT-2, and BART \citep{devlin2018bert,radford2019language,lewis2019bart}, trained on extensive corpora to acquire universal language representations and fine-tuned for specific downstream tasks, such as text classification or question answering.

The success of PLMs prompted extensive research, particularly in scaling PLMs. Large-scale PLMs such as GPT-3 \citep{brown2020language}, GPT-4 \citep{openai2023gpt4}, PaLM \citep{chowdhery2022palm}, and Llama \citep{chowdhery2022palm} demonstrated remarkable capabilities in complex tasks, leading to the term "large language model (LLM)." Notable applications of LLMs include ChatGPT \citep{openai_chatgpt} for dialogue interaction and Med-PaLM 2 \citep{singhal2023towards} for medical question answering.

A distinct characteristic of LLMs is their 'knowledge emergence' ability  \citep{wei2022emergent}, absent in smaller models but present in larger ones~\citep{brown2020language,wei2021finetuned,wei2022chain}. This ability manifests in scenarios like few-shot prompting \citep{brown2020language}, where LLMs generate desired outputs with minimal demonstrations and natural language instructions, without further training. Instruction tuning \citep{wei2021finetuned} enhances LLMs' task adaptability by fine-tuning on a mix of tasks presented as instructions. Chain-of-thought (CoT) prompting \citep{wei2022chain}, an advanced strategy, involves incorporating intermediate reasoning steps into prompts to improve LLM performance in complex reasoning tasks.

The development of LLMs has not only revolutionized the field of NLP but also holds significant potential for other disciplines. Consequently, LLMs are likely to remain a central focus of research and development. \tablename~\ref{table-1} offers a concise comparison of traditional statistical models, neural language models, PLMs, and LLMs.

\begin{table*}[h]
    \centering   
    \caption{Comparison of traditional statistical Models, neural language models, pre-trained language models and large language models.}
    \label{table-1}
    \resizebox{18cm}{!}{
    \begin{tabular}{c c c c c}
    \toprule
        \multicolumn{1}{c}{\textbf{Comparison}} & \multicolumn{1}{c}{\textbf{Traditional Statistical Models}} & \multicolumn{1}{c}{\textbf{Neural Language Models}} & \multicolumn{1}{c}{\textbf{Pre-trained Language Models}} & \multicolumn{1}{c}{\textbf{Large Language Models}} \\
        \midrule
        \textbf{Model Size} & Limited & Large & Large & Very large \\
        \textbf{Training Data Size} & Large & Large & Large & Very large \\
        \textbf{Emergent Abilities} & No & No &	No & Yes \\
        \textbf{Feature Extraction} & Artificial & Automatic & Automatic & Automatic \\
        \textbf{Interactiveness} & Poor & Poor & Poor & Good \\
        \textbf{Interpretability} & Good & Poor & Poor & Poorest \\
        \textbf{Performance} & Common & Higher & Higher & Highest \\
        \textbf{Evaluation} & Automatic	& Automatic	& Automatic & Automatic,
        Human \\
        \textbf{Resources Requirements} & Low & High & High & Highest \\
        \textbf{Ability to Capture Long-range Dependencies} & Poor & Better & Better & Best \\
        \textbf{Representative Models} & N-gram & RNN,LSTM & BERT,GPT-1,T5 & PaLM,GPT-3,Llama,GPT-4 \\
        \bottomrule
    \end{tabular}}
\end{table*}

\subsection{Diffusion Models}

Diffusion models, a class of probabilistic generative models in machine learning and image processing, have garnered attention for their unique approach to data evolution over time through controlled, incremental diffusion steps \citep{sohl2015deep,ho2020denoising,song2019generative,song2020score}. These models start with a real image and progressively introduce Gaussian noise at each step, transforming the image into a progressively noisier version. The training process involves reversing this noise-addition, effectively restoring the image to its original state.

Given an original image $x_{0}$, the forward process of a diffusion model incrementally adds Gaussian noise over $T$ steps, resembling a Markov process where each time $t$ depends solely on the preceding $t-1$. This process gradually generates the latent variables $x_{1},x_{2},\ldots,x_{T}$. The forward process's probability distribution $q(\cdot)$ is generated as follows:
\[q(x_{1:T}
|x_{0})=\prod_{i=1}^{T}q(x_{t}|x_{t-1})\tag{1}\]
Each transformation step is defined as a Gaussian transformation $\mathcal{N}(\cdot)$ following a schedule $\{\beta_{t}\in(0,1)\}_{t=1}^{T}$:
\[q(x_{t}|x_{t-1})=\mathcal{N}(x_{t};\sqrt{1-\beta_{t}}x_{t-1},\beta_{t}I)\tag{2}\]
As $t$ progresses, $x_{t}$ gradually approaches pure noise.

Conversely, the reverse process represents denoising inference, training a neural network to sequentially remove noise from an entirely noisy image to recover the real image.
\[p_{\theta}(x_{0:T})=p(x_{T})\prod_{t=1}^{T}p_{\theta}(x_{t-1}|x_{t})\tag{3}\]

The transformation $p(\cdot)$ is designed to progressively reduce variance. Consequently, the ultimate sample $x_{0}$ signifies a sample extracted from the true distribution. This transformation is typically parameterized by a fixed covariance $\sum_{t}=\beta_{t}I$ and a learning mean $\mu_{\theta}(x_{t},t)$ as defined below:
\[\mu_{\theta}(x_{t},t)=\frac{1}{\sqrt{\alpha_{t}}}(x_{t}-\frac{\beta_{t}}{\sqrt{1-\overline{\alpha}_{t}}}\epsilon_{\theta}(x_{t},t))\tag{4}\]
Where $\epsilon_{\theta}(x_{t},t)$ denotes a trained neural network tasked with processing noisy samples $x_{t}$ and predicting the introduced noise. Given an authentic sample $x_{0}$ and noise $\epsilon\sim N(0,1)$, $x_{t}$ can be computed at any given time step according to the following:
\[x_{t}(x_{0},\epsilon)=\sqrt{\overline{\alpha}_{t}}x_{0}+\sqrt{1-\overline{\alpha}_{t}}\epsilon\tag{5}\]
Where $\alpha_{t}=1-\beta_{t}$and $\overline{\alpha}_{t} = \prod_{s=1}^{t}\alpha_{t}$.

In multimodal scenarios, combining image and language offers comprehensive insights. Allowing concurrent guidance from both image and language introduces extra flexibility, providing users with enhanced control options \citep{liu2023more}. Using diffusion models for multimodal data augmentation involves configuring additional modes as conditions, particularly through a denoising network $\epsilon_{\theta}(x_{t},y,t)$ modulated by an auxiliary input $y$. This allows for sampling from a data distribution conditioned on $y$. With pre-training integration, diffusion models excel in applications like image editing \citep{ruiz2023dreambooth,kawar2023imagic,couairon2022diffedit,brooks2023instructpix2pix} and image inpainting \citep{xie2023smartbrush,nichol2021glide} based on text prompts, showcasing their versatility in enhancing multimodal data.

\subsection{Data Augmentation}

Data augmentation, a cornerstone technique in deep learning, becomes vital when faced with data collection challenges. By employing diverse augmentation strategies, it enriches datasets, expanding their scope, enhancing the training model's robustness, and refining its generalization capabilities.

A primary function of data augmentation is to counter overfitting, often encountered in training deep neural networks \citep{shorten2021text}. Without augmentation or regularization, these networks risk adopting spurious correlations and memorizing intricate, high-frequency patterns that may elude human detection.

In fields like computer vision (CV) and natural language processing (NLP), data augmentation plays a critical role in improving model robustness and generalization. In CV, traditional methods include random rotation, flipping, scaling, and cropping, introducing variations in orientation and scale. Additional techniques like color dithering and noise addition further diversify the dataset. Innovative methods like image hybrid data enhancement, exemplified by Mixup \citep{zhang2017mixup} and CutMix \citep{yun2019cutmix}, blend images or their sub-regions, enhancing data diversity beyond basic image processing techniques. In NLP, Easy Data Augmentation (EDA) \citep{wei2019eda} techniques like synonym replacement, random insertion, random swap, and random deletion are prevalent. However, these methods face limitations, such as the need for the available data's distribution to closely mirror the actual data distribution, risks of information loss or distortion, and challenges in maintaining labeling consistency.

Adversarial training \citep{goodfellow2014explaining,miyato2016adversarial,madry2017towards,shafahi2019adversarial} is a technique where models are exposed to adversarial examples during training. These examples are crafted to deceive the model, forcing it to become more robust against potential adversarial attacks. This approach enhances the model's resilience by teaching it to handle unexpected variations and disturbances in the input data.

Generative modeling, another promising avenue for data enhancement, creates artificial instances that retain features similar to the original dataset. Generative Adversarial Networks (GANs) \citep{goodfellow2020generative}, particularly influential in CV, have evolved into various forms, becoming a robust tool for dataset enhancement \citep{antoniou2017data,brock2018large,xia2018auxiliary}.

\begin{figure*}[htbp]
    \centering
    \includegraphics[width=\textwidth]{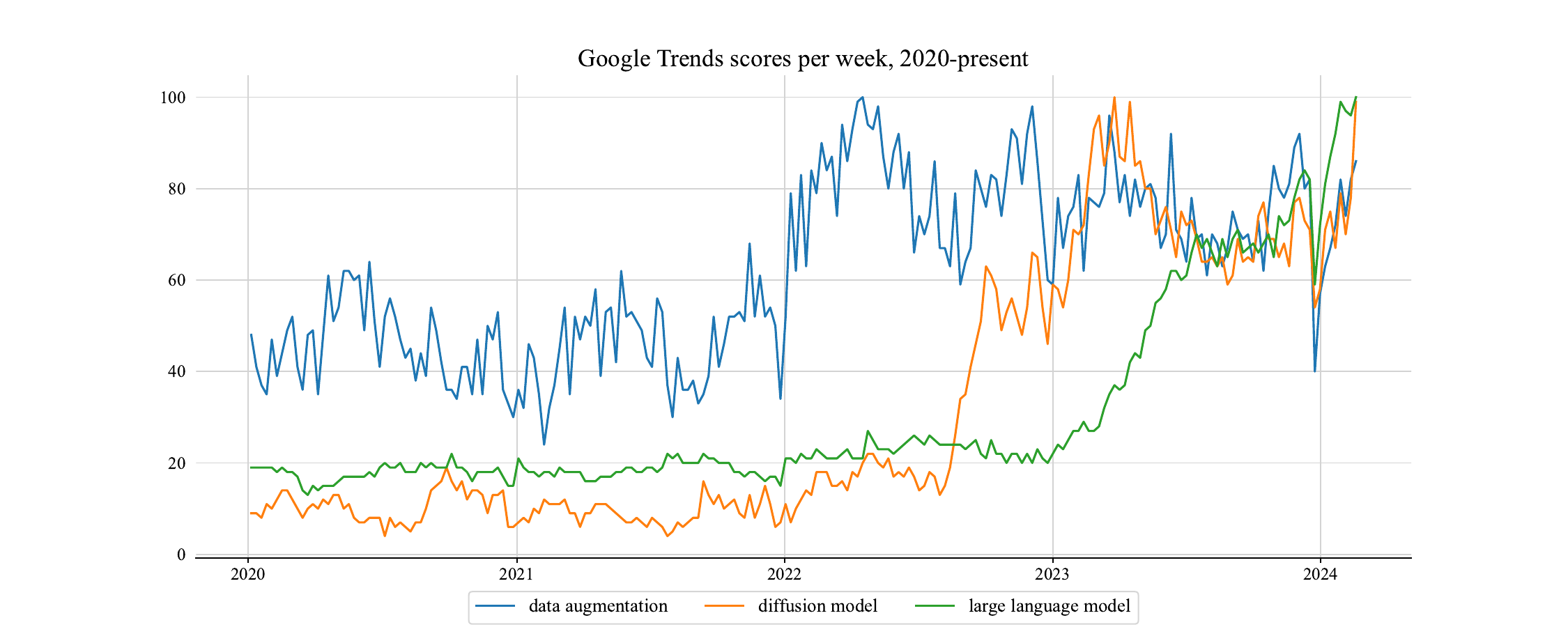}
    \caption{Weekly Google Trends scores for the search term "large language model", "diffusion model", and "data augmentation".}
    \label{fig:trend}
\end{figure*}

Recent advancements in large models like GPT-3, T5 \citep{raffel2020exploring}, Stable Diffusion \citep{rombach2022high}, and CLIP \citep{radford2021learning} have shown exceptional performance in data augmentation. Pre-trained on extensive datasets and fine-tuned for specific tasks, large models offer rich representation patterns, serving as a robust foundation for various downstream tasks. Initially designed for NLP, LLMs' versatility extends to image-related challenges, introducing context-aware transformations and enhancing semantic features beyond traditional methods' capabilities \citep{radford2021learning,nichol2021glide,zhang2023adding}. 
This paper endeavors to sensitize the research community towards the burgeoning interest in large models, data augmentations, and their combinations (as seen in \figurename~\ref{fig:trend}). \tablename~\ref{tab:comparisonDataAug} delineates the differences between large model-driven and traditional data augmentation methods across various aspects.

\begin{table*}[h]
    \centering
    
    \caption{Comparison of large model-driven and traditional methods for data augmentation.}
    \label{tab:comparisonDataAug}
    \begin{tabular}{>{\centering}m{2cm}>{\raggedright}m{4.5cm}>{\raggedright\arraybackslash}m{4.5cm}}
       \toprule
            \multicolumn{1}{c}{\textbf{Aspect}} & \multicolumn{1}{c}{\textbf{Large Model-Driven Methods}} & \multicolumn{1}{c}{\textbf{Traditional Methods}} \\
            \midrule
            \textbf{Semantic Consistency} & Contributes to higher semantic consistency by leveraging language understanding. & Lack of explicit mechanisms for maintaining semantic consistency. \\
            \midrule
            \textbf{Flexibility and Creativity} & Demonstrates high flexibility and creativity in generating diverse content. & May be constrained by predefined transformation patterns, limiting creative variations. \\
            \midrule
            \textbf{Data Diversity} & Learns diverse contexts from extensive data, introducing varied augmentation. & May rely on limited data sources, potentially resulting in less diverse augmentation. \\
            \midrule
            \textbf{Computational Efficiency} & Exhibits lower computational efficiency due to a large number of parameters. & Tends to be more computationally efficient, especially in basic geometric transformations. \\
            \midrule
            \textbf{Domain Specificity} & Domain-agnostic, applicable to various text and image data types. & Traditional methods may excel in specific domains, optimized for domain-specific features. \\
            \bottomrule
    
    \end{tabular}
\end{table*}

\section{Approaches}
\label{sec-appro}

The advent of large models has revolutionized data augmentation, offering novel and effective means to generate training data with greater diversity compared to traditional methods. This section categorizes existing methodologies into three distinct classes based on the target data type: image augmentation, text augmentation, and paired data augmentation. Image augmentation pertains to expanding image data, text augmentation to expanding text data, and paired data augmentation to both. These methods reflect the latest trends in data augmentation, highlighting the significant role of large models.

\subsection{Image Augmentation}

Image augmentation synthesizes realistic images, guided by additional information. We divide these techniques into prompt-driven and subject-driven approaches: text, visual, and multimodal approaches in the prompt-driven category; and subject-specific strategies in the subject-driven category. Text prompt-driven approaches generate images from textual descriptions, visual prompt-driven approaches use visual cues, and multimodal prompt-driven approaches combine both textual descriptions and visual guidance. Subject-driven approaches tailor augmentation for specific subjects. These approaches enhance deep learning task performance, contributing to more robust training experiences. Existing approaches are summarized in Table \ref{table:imgAug}.

\begin{table*}[]
\centering
\caption{Summary of image augmentation including prompt-driven and subject-driven approaches.}
\label{table:imgAug}
\begin{adjustbox}{width=\textwidth}
\begin{tabular}{|l|ccc|c|}
\hline
\multirow{2}{*}{\textbf{Reference}} & \multicolumn{3}{c|}{\textbf{Prompt-driven approaches}}& \multirow{2}{*}{\textbf{Subject-driven approaches}} \\ \cline{2-4}
                           & \multicolumn{1}{c|}{\textbf{Text prompt-driven}} & \multicolumn{1}{c|}{\textbf{Visual prompt-driven}} & \textbf{Multimodal prompt-driven} &\\ \hline

\citet{avrahami2022blended}& \multicolumn{1}{c|}{\cmark} & \multicolumn{1}{c|}{} & & \\ \hline
\citet{bar2022text2live}& \multicolumn{1}{c|}{\cmark} & \multicolumn{1}{c|}{} & & \\ \hline
\citet{brooks2023instructpix2pix}& \multicolumn{1}{c|}{\cmark} & \multicolumn{1}{c|}{} & & \\ \hline

\citet{chen2023subject}& \multicolumn{1}{c|}{} & \multicolumn{1}{c|}{} & &\cmark \\ \hline
\citet{couairon2022diffedit}& \multicolumn{1}{c|}{\cmark} & \multicolumn{1}{c|}{} & & \\ \hline
\citet{doubinsky2023semantic}& \multicolumn{1}{c|}{\cmark} & \multicolumn{1}{c|}{} & & \\ \hline
\citet{du2023boosting}& \multicolumn{1}{c|}{} & \multicolumn{1}{c|}{} &\cmark & \\ \hline
\citet{dunlap2023diversify}& \multicolumn{1}{c|}{\cmark} & \multicolumn{1}{c|}{} & & \\ \hline
\citet{erfanian2024chameleon}& \multicolumn{1}{c|}{} & \multicolumn{1}{c|}{} &\cmark & \\ \hline
\citet{feng2024instagen}& \multicolumn{1}{c|}{\cmark} & \multicolumn{1}{c|}{} & & \\ \hline
\citet{gal2022image}& \multicolumn{1}{c|}{} & \multicolumn{1}{c|}{} & &\cmark \\ \hline
\citet{gal2022stylegan}& \multicolumn{1}{c|}{\cmark} & \multicolumn{1}{c|}{} & & \\ \hline
\citet{ge2023expressive}& \multicolumn{1}{c|}{\cmark} & \multicolumn{1}{c|}{} & & \\ \hline
\citet{hemati2023cross}& \multicolumn{1}{c|}{\cmark} & \multicolumn{1}{c|}{} & & \\ \hline
\citet{hertz2022prompt}& \multicolumn{1}{c|}{\cmark} & \multicolumn{1}{c|}{} & & \\ \hline
\citet{huang2023reversion}& \multicolumn{1}{c|}{} & \multicolumn{1}{c|}{} &\cmark & \\ \hline
\citet{kawar2023imagic}& \multicolumn{1}{c|}{\cmark} & \multicolumn{1}{c|}{} & & \\ \hline
\citet{kim2022diffusionclip}& \multicolumn{1}{c|}{\cmark} & \multicolumn{1}{c|}{} & & \\ \hline
\citet{koohpayegani2023genie}& \multicolumn{1}{c|}{\cmark} & \multicolumn{1}{c|}{} & & \\ \hline
\citet{kumari2023multi}& \multicolumn{1}{c|}{} & \multicolumn{1}{c|}{} & &\cmark \\ \hline
\citet{li2023blip}& \multicolumn{1}{c|}{} & \multicolumn{1}{c|}{} & &\cmark \\ \hline
\citet{li2023gligen}& \multicolumn{1}{c|}{\cmark} & \multicolumn{1}{c|}{} & & \\ \hline
\citet{liu2023more}& \multicolumn{1}{c|}{} & \multicolumn{1}{c|}{\cmark} & & \\ \hline
\citet{luo2023camdiff}& \multicolumn{1}{c|}{\cmark} & \multicolumn{1}{c|}{} & & \\ \hline
\citet{ma2023unified}& \multicolumn{1}{c|}{} & \multicolumn{1}{c|}{} & &\cmark \\ \hline
\citet{nguyen2023visual}& \multicolumn{1}{c|}{} & \multicolumn{1}{c|}{} &\cmark & \\ \hline
\citet{nichol2021glide}& \multicolumn{1}{c|}{\cmark} & \multicolumn{1}{c|}{} & & \\ \hline
\citet{patashnik2023localizing}& \multicolumn{1}{c|}{\cmark} & \multicolumn{1}{c|}{} & & \\ \hline
\citet{ruiz2023dreambooth}& \multicolumn{1}{c|}{} & \multicolumn{1}{c|}{} & &\cmark \\ \hline
\citet{samuel2023all}& \multicolumn{1}{c|}{\cmark} & \multicolumn{1}{c|}{} & & \\ \hline
\citet{schnell2023generative}& \multicolumn{1}{c|}{} & \multicolumn{1}{c|}{} &\cmark & \\ \hline
\citet{shi2023instantbooth} & \multicolumn{1}{c|}{} & \multicolumn{1}{c|}{} & &\cmark \\ \hline
\citet{sun2023imagebrush}& \multicolumn{1}{c|}{} & \multicolumn{1}{c|}{\cmark} & & \\ \hline
\citet{trabucco2023effective}& \multicolumn{1}{c|}{\cmark} & \multicolumn{1}{c|}{} & & \\ \hline
\citet{tumanyan2023plug}& \multicolumn{1}{c|}{\cmark} & \multicolumn{1}{c|}{} & & \\ \hline
\citet{wang2023context}& \multicolumn{1}{c|}{} & \multicolumn{1}{c|}{} &\cmark & \\ \hline
\citet{wang2023dynamic}& \multicolumn{1}{c|}{\cmark} & \multicolumn{1}{c|}{} & & \\ \hline
\citet{wang2024diffusion}& \multicolumn{1}{c|}{} & \multicolumn{1}{c|}{} &\cmark & \\ \hline
\citet{wei2023elite}& \multicolumn{1}{c|}{} & \multicolumn{1}{c|}{} & &\cmark \\ \hline
\citet{wu2023image}& \multicolumn{1}{c|}{} & \multicolumn{1}{c|}{\cmark} & & \\ \hline
\citet{xiao2023fastcomposer}& \multicolumn{1}{c|}{} & \multicolumn{1}{c|}{} & &\cmark \\ \hline
\citet{xie2023smartbrush}& \multicolumn{1}{c|}{} & \multicolumn{1}{c|}{} &\cmark & \\ \hline
\citet{yin2023ttida}& \multicolumn{1}{c|}{\cmark} & \multicolumn{1}{c|}{} & & \\ \hline
\citet{yu2023diffusion}& \multicolumn{1}{c|}{} & \multicolumn{1}{c|}{\cmark} & & \\ \hline
\citet{zhang2023sine}& \multicolumn{1}{c|}{\cmark} & \multicolumn{1}{c|}{} & & \\ \hline

\end{tabular}
\end{adjustbox}
\end{table*}

\subsubsection{Prompt-driven approaches}

\textbf{Text prompt-driven approaches} 
CamDiff \citep{luo2023camdiff} employs the latent diffusion model (LDM) \citep{rombach2022high} to synthesize salient objects in camouflage scenes, leveraging CLIP's zero-shot image classification capabilities to prevent synthesis failures and maintain consistency with input prompts. 
\citet{couairon2022diffedit} proposed DiffEdit, an algorithm that automatically finds what regions of an input image should be edited based on a textual query.
GLIDE \citep{nichol2021glide} investigates diffusion models for text-driven image synthesis and compares two guidance approaches: CLIP-guided and classifier-free guidance. Classifier-free guidance is preferred by evaluators for its photorealism and caption similarity capabilities, resulting in highly realistic output. The authors also demonstrated the potential of powerful text-driven image editing by fine-tuning models for image inpainting.
SeedSelect \citep{samuel2023all} is an innovative method for refining the generation of unconventional and poorly-formed concepts in diffusion models. This technique optimizes the generation process by identifying appropriate generation seeds using a small reference set of images, ensuring the accurate generation of rare concepts both semantically and visually.

\citet{tumanyan2023plug} presented a framework for image-to-image translation, employing a pre-trained text-to-image diffusion model to generate images based on guidance and text prompts. The model offers precise control over structure through spatial feature manipulation, simplifying the process without additional training.
\citet{hertz2022prompt} presented a prompt-to-prompt editing framework that is controlled by text. It focuses on analyzing a text-conditioned model and highlights the importance of cross-attention layers in managing the relationship between image spatial layout and prompt.
\citet{patashnik2023localizing} presented a technique for generating diverse images showcasing variations in the shape of a specific object, facilitating object-level exploration. In order to control the object's shape while preserving semantics, the specific challenge of accurate shape manipulation localization is addressed by introducing a prompt-mixing technique during denoising. Additionally,  two techniques using self-attention and cross-attention layers are proposed to pinpoint image-space operations. 
SINE \citep{zhang2023sine} is a single-image editing method that relies on a pre-trained text-to-image diffusion model. By fine-tuning this model with a single image and a brief textual description, it enables versatile image editing at any resolution, maintaining both fidelity and generalization. 

Text2LIVE \citep{bar2022text2live} is a model for semantically meaningful object appearance editing and visual effects. It generates an RGBA edit layer composited over the input based on text prompts, enabling precise content control through text-driven objectives.
DiffusionCLIP \citep{kim2022diffusionclip} is a text-guided image manipulation method that utilizes pre-trained diffusion models and CLIP loss. It excels in in-domain and out-of-domain manipulation after fine-tuning. 
StyleGAN-NADA \citep{gal2022stylegan} combines StyleGAN and CLIP to transfer generative models to new domains in a text-driven manner without collecting any images.
\citet{dunlap2023diversify} presented automated language-guided image augmentation (ALIA), a novel approach that utilizes large vision and language models to automatically generate natural language descriptions of a dataset’s domains and augment the training data via language-guided image editing. As a result, the augmented dataset maintains visual consistency with the original training data while offering considerably increased diversity. 
\citet{trabucco2023effective} introduced DA-Fusion, data augmentation by fusion, a flexible data augmentation strategy that employs text-to-image diffusion models to generate diverse variations of real images. This method facilitates semantic editing of images through an off-the-shelf diffusion model, allowing for the modification of image semantics. Furthermore, DA-Fusion showcases its proficiency in adapting to novel visual concepts with a scant amount of labeled instances, surmounting the constraints of conventional data augmentation methods that predominantly depend on rudimentary transformations like rotations and flips, thereby offering limited semantic variance in the creation of new images from pre-existing ones.

\citet{kawar2023imagic} presented Imagic, a text-conditioned image editing method that utilizes a pre-trained text-to-image diffusion model. Imagic enables image editing by simply providing a single input image and a target text, eliminating the need for additional inputs such as image masks or additional views of the object. 
\citet{yin2023ttida} introduced text-to-text-to-image data augmentation (TTIDA), an innovative method that combines the capabilities of large-scale pre-trained text-to-text models (GPT-2) with text-to-image models (GLIDE) to perform data augmentation. TTIDA leverages both models to generate diverse and realistic textual descriptions and corresponding images, enhancing the dataset for training.
\citet{avrahami2022blended} proposed the first solution for general-purpose local image editing. This approach utilizes a natural language description and an ROI mask as guidance. Combining a pre-trained vision-language model (CLIP) with a denoising diffusion probabilistic model (DDPM), the proposed method generates results that exhibit a natural and realistic appearance. 
\citet{wang2023dynamic} introduced dynamic prompt learning (DPL) as a technique to address the issue of background and distractor object leakage in image editing with text-to-image diffusion models. It accomplished this by dynamically updating the tokens associated with each noun word in the prompt. This adjustment minimizes attention leakage in the cross-attention maps, leading to better results in text-guided image editing.
\citet{brooks2023instructpix2pix} introduced a method for image editing guided by human instructions, combining knowledge from pre-trained language and text-to-image models to create a large training dataset, the design of the InstructPix2Pix enables fast image editing by efficiently generalizing to real images and user-written instructions during inference without fine-tuning or inversion. 

\citet{ge2023expressive} addressed the limitations of plain text in text-to-image synthesis by introducing a rich-text editor that supports attributes such as font style, size, color, and footnotes. This allows for precise customization and fine-grained control. A region-based diffusion process is employed to ensure the fidelity of the generated images, permitting detailed prompts and region-specific guidance.
Recognizing the effectiveness and efficiency of augmented samples near a classifier's ideal decision boundary, GeNIe \citep{koohpayegani2023genie} is introduced. GeNIe utilizes a diffusion model conditioned on a text prompt to merge divergent data points (an image from the source category and a text prompt from the target category), generating challenging samples for the target category. Inspired by contemporary image editing methods, the model regulates both the number of diffusion iterations and the noise level. This regulation guarantees the preservation of low-level and contextual features from the source image in the generated image, potentially conflicting with the target category.

\citet{doubinsky2023semantic} investigated the enhancement of few-shot class-agnostic counting using synthetic data. A dual conditioning approach, utilizing Stable Diffusion with both a prompt and a density map, is proposed to augment a small training dataset for few-shot counting. To overcome limited dataset diversity, a strategy is introduced involving caption swapping between images and creating novel object configurations and spatial layouts.
\citet{feng2024instagen} presented InstaGen, an innovative method for training object detection models using synthetic datasets created by fine-tuning generative diffusion models. InstaGen boosts detection accuracy by integrating instance-level bounding boxes into synthetic images, facilitating training across an expanded set of categories, and enhancing performance robustness in practical applications.
\citet{hemati2023cross} introduced Prompt-guided cross domain generative augmentation (CDGA-PG), which involves taking an image from a specific domain and class and then utilizing text descriptions from other domains as prompts. This process aims to enhance the original image by transforming it into versions that belong to the same class but appear as if they come from other domains. This method bridges the gap between domains, reduces distribution shifts, and improves model generalization in domain generalization tasks.

\noindent \textbf{Visual prompt-driven approaches}
ImageBrush \citep{sun2023imagebrush} is designed for precise image editing guided by visual instructions. It utilizes transformation images that are extracted from visual demonstrations as instructions and employs a diffusion-based inpainting approach to uncover human intent, enhancing the model's capacity for accurate editing. 
\citet{yu2023diffusion} introduced a diffusion-based method for augmenting nuclei segmentation datasets. It employs a two-step process: first, it generates synthetic nuclei structures; second, it uses these structures to synthesize histopathology images. The resulting synthetic images closely mimic real samples, align well with the nuclei structures, and exhibit diverse styles, showcasing a promising potential for segmentation model training.

\citet{wu2023image} enhanced weakly-supervised semantic segmentation (WSSS) by using image augmentation with controlled diffusion (IACD) method. This technique enriches labeled datasets by utilizing available images, image-level labels, and detection maps as control inputs. Furthermore, it implemented a robust image selection strategy to mitigate noise in the diffusion model. The experimental results demonstrate that the IACD approach surpasses existing methods in performance, particularly in scenarios with limited data availability, thereby underscoring its effectiveness.
Semantic diffusion guidance (SDG) \citep{liu2023more} employs a fine-tuned image encoder for Image-guided image synthesis. Similar to language guidance, it extracts embeddings to capture high-level semantics. The use of image encoders allows control over retaining structural information from the reference image, despite embeddings lacking spatial dimensions. By leveraging spatial feature maps and enforcing alignment, SDG guides generated images to share similar structures with the reference image. Through image-guided diffusion, it achieves diverse image synthesis aligned with the semantics of the guidance image.

\noindent \textbf{Multimodal prompt-driven approaches}
\citet{nguyen2023visual} introduced a framework for image editing via visual prompt inversion. With just one example pair illustrating the "before" and "after" states of an image editing task, this approach achieves competitive results compared to state-of-the-art text-conditioned image editing models.
Prompt Diffusion \citep{wang2023context} is a framework for enabling in-context learning within diffusion-based generative models. Given pairs of task-specific images and text guidance, this model automatically comprehends and replicates tasks on new images. It introduces a versatile vision-language prompt to model various tasks and is jointly trained on six tasks, demonstrating high-quality in-context generation and effective task generalization, including text-guided image editing.
SmartBrush \citep{xie2023smartbrush} introduces precise content completion in missing regions by combining text and visual guidance from masks. The model incorporates innovative training and sampling techniques, including object-mask prediction, to enhance background preservation. Additionally, a multi-task training approach is applied, jointly training inpainting and text-to-image generation to maximize the use of extensive training data.

ReVersion \citep{huang2023reversion} tackles the task of relation inversion, concentrating on acquiring distinct relations through the use of relation prompts derived from exemplary images. This method employs relation prompts, which are formulated utilizing Stable Diffusion, to synthesize images that are specific to a particular relation, incorporating novel elements. The technique underscores the concept of a "prepositional prior," wherein prompts based on real-world relations are selectively activated by leveraging fundamental prepositional words. This is accomplished through the introduction of an innovative relation-steering contrastive learning framework, designed to capture the dynamics of object interactions while concurrently segregating the appearances of objects. Furthermore, the method employs relation-focused importance sampling to accentuate high-level interactions, thus prioritizing them over minutiae.
GLIGEN \citep{li2023gligen} enhances LDM by integrating new layers into its existing structure. This advancement enables more effective grounded language-to-image generation. The model adeptly generates images based on inputs such as captions and bounding boxes, exhibiting strong adaptability to new spatial configurations and concepts. The method is both simple and efficient, offering the flexibility to extend to other conditions like keypoints, reference images, and various spatially-aligned conditions, including edge and depth maps.

ControlNet \citep{zhang2023adding}, a neural network architecture, adds spatial conditioning controls to large, pre-trained text-to-image diffusion models. It harnesses the established diffusion models' deep encoding layers, trained on billions of images, as a base for learning varied conditional controls. The architecture features "zero convolutions" - convolution layers initialized at zero - which progressively increase parameters, ensuring fine-tuning is unaffected by harmful noise.
Drawing from ControlNet, \citet{du2023boosting} employed lesion-specific visual and textual prompts to generate dermatoscopic images. The framework integrates a controllable lesion function, enabling manipulation of lesion type, textual attributes, and shapes with corresponding locations in mask images during both training and inference. The learned correlation between visual and textual prompts prioritizes rare cases during inference. Additionally, an automated module is introduced for generating lesion shapes and masks. 
\citet{schnell2023generative} proposed a method that utilizes a ControlNet diffusion model, conditioned on semantic scribbles, for generating high-quality training data. To ensure class consistency, the model employs classifier-free diffusion guidance. Encode ratios are introduced to balance data diversity and realism. The proposed augmentation schemes, influenced by guidance scale and encode ratio, yield a spectrum of high-quality training images.

Chameleon \citep{erfanian2024chameleon} is a system that leverages foundation models (e.g., DALL·E 2 \citep{OpenAI2022DALLe2}) for fairness-aware multi-modal data augmentation to enhance coverage of minorities, addressing under-representation in training data and reducing unfairness in downstream tasks. To ensure the model-generated images align with the original dataset's distribution, this process encompasses specifying a text prompt as well as incorporating a guide tuple and mask.
\citet{wang2024diffusion} introduced a novel framework that leverages diffusion models (ControlNet) to generate synthetic datasets tailored for crowd-counting tasks. This method enhances the precision of crowd-counting models by creating an extensive array of training data, encompassing varied scenes and crowd densities. This diversity is achieved by modifying the conditioned text and location prompts.

\subsubsection{Subject-driven approaches}

Subject-driven approaches aim at synthesizing diverse and personalized images based on user-provided images capturing a specific subject. In contrast to general image generation techniques, subject-driven generation focuses on allowing users to create novel renditions of a subject in various contexts while maintaining its distinctive features. This approach caters to the user's desire for customized and imaginative outputs. It is especially noteworthy that subject-driven approaches can generate a variety of images with diverse backgrounds from a small number of subject images.

\citet{gal2022image} introduced a method, utilizing LDM, for generating novel "words" within a text-to-image model's embedding space with the aid of 3-5 conceptual images, facilitating intuitive, personalized image creation guided by language.
\citet{ruiz2023dreambooth} proposed a method called DreamBooth, which can generate a variety of photorealistic images of a subject in different contexts with a few reference images and a text prompt. This method has the capacity to create innovative versions of the subject in diverse situations while preserving its unique characteristics. 
InstantBooth \citep{shi2023instantbooth} allows instant text-guided image personalization without test-time fine-tuning. This is achieved by converting images into textual tokens for concept learning and incorporating adapter layers to preserve fine details and identity during image generation, without the use of paired images of the same concept.

UMM-Diffusion \citep{ma2023unified} is a method for generating customized images by encoding text and images jointly into a unified multimodal latent space. This method combines text and image information to guide image generation and eliminates irrelevant image parts through a novel sampling technique. 
Custom Diffusion \citep{kumari2023multi} is a fast and efficient fine-tuning technique for text-to-image diffusion models that updates key and value mapping weights in cross-attention layers for new concepts, uses real images with similar captions to prevent forgetting, and introduces augmentation for faster convergence. It also supports training of multiple concepts together or separately and merging.
BLIP-Diffusion \citep{li2023blip} is a subject-driven image generation model with multimodal control using subject images and text prompts. It utilizes a pre-trained multimodal encoder, aligned visual representation with text following BLIP-2, and a subject representation learning task to generate new subject renditions.

FastComposer \citep{xiao2023fastcomposer} presents a tuning-free technique for generating personalized, multi-subject text-to-image content. By utilizing a pre-trained vision encoder, this approach effectively tackles identity blending concerns by training supervised cross-attention maps with segmentation masks. 
ELITE \citep{wei2023elite} is a learning-based encoder specifically crafted for rapid and precise customized text-to-image generation. It introduces two key components: a global mapping network that translates image features into "new" text, and a local mapping network dedicated to preserving details and concept editability. 
SuTI \citep{chen2023subject} introduces a subject-driven text-to-image generation framework capable of producing a wide array of images depicting a given subject across various scenes, eliminating the necessity for subject-specific fine-tuning. This framework adopts an apprenticeship learning strategy, wherein it acquires knowledge from an extensive compilation of subject-specific expert models, thereby enabling the generation of nuanced and contextually rich images.

\subsection{Text Augmentation}
Text augmentation focuses on harnessing the advanced capabilities of large models to augment text datasets, which includes two strategies: label-based and generated content-based. In the label-based approach, models are employed to annotate text data, effectively enriching the text dataset with a larger volume of labeled instances. Generated content-based strategies guide models to synthesize new text data to expand the dataset with freshly generated textual material. The existing methods are shown in \tablename~\ref{table:texaug}.

\begin{table*}[ht!]
\centering
\caption{Summary of text augmentation including label-based and generated content-based approaches.}
\label{table:texaug}
    \begin{tabular}{|m{4cm}|c|c|c|c|c|c|c|c|c|c|}
    \hline
    Reference & Label-based & Generated content-based \\ \hline
 
\citet{bonifacio2022inpars} & & \cmark \\ \hline
\citet{cai2023resolving} & & \cmark \\ \hline
\citet{chen2023minprompt} & \cmark & \\ \hline
\citet{chowdhury2023generative} & \cmark & \\ \hline
\citet{dai2023chataug} & & \cmark \\ \hline
\citet{gao2022self} & & \cmark \\ \hline
\citet{guo2022genius} & & \cmark \\ \hline
\citet{jo2022dagam} & & \cmark \\ \hline
\citet{kaddour2023text} & \cmark & \\ \hline
\citet{khatri2022skillbot} & & \cmark \\ \hline
\citet{kumar2020data} & \cmark & \cmark \\ \hline
\citet{latif2023can} & \cmark & \\ \hline
\citet{li2023dail} & & \cmark \\ \hline
\citet{liu2022wanli} & & \cmark \\ \hline
\citet{lu2023epa} & & \cmark \\ \hline
\citet{meng2023tuning} & & \cmark \\ \hline
\citet{oh2023data} & & \cmark \\ \hline
\citet{quteineh2020textual} & & \cmark \\ \hline
\citet{saakyan2023iclef} & & \cmark \\ \hline
\citet{samuel2023can} & \cmark & \\ \hline
\citet{sahu2022data} & \cmark & \cmark \\ \hline
\citet{sachdeva2023catfood} & & \cmark \\ \hline
\citet{schlegel2023pulsar} & & \cmark \\ \hline
\citet{sharma2023augmenting} & \cmark & \cmark \\ \hline
\citet{tang2023just} & & \cmark \\ \hline
\citet{tarjan2020deep} & & \cmark \\ \hline
\citet{thakur2020augmented} & \cmark & \\ \hline
\citet{wang2024large}& & \cmark \\ \hline
\citet{wu2023improving} & \cmark & \\ \hline
\citet{ye2024llm}& \cmark & \\ \hline
\citet{yoo2021gpt3mix} & \cmark & \cmark \\ \hline
\citet{yu2022coca} & & \cmark \\ \hline
\citet{zhou2021flipda} & & \cmark \\ \hline
\citet{zheng2023augesc} & & \cmark \\ \hline

    \end{tabular}
\end{table*}

\subsubsection{Label-based approaches}
\citet{thakur2020augmented} introduced an effective data augmentation approach called Augmented SBERT. This method utilizes the cross-encoder BERT to annotate a larger collection of input pairs, thereby enhancing the training data for the bi-encoder SBERT model. \citet{kumar2020data} investigated the utilization of pre-trained models like GPT-2, BERT, and BART \citep{lewis2019bart} for data augmentation with the aim of enhancing text classification accuracy. \citet{sahu2022data} presented a prompting-based strategy to produce labeled training data for intent classification using general-purpose language models (LMs) like GPT-3. This method offers the advantage of not needing task-specific LM fine-tuning for data generation, eliminating the need for hyperparameter tuning. Additionally, this approach remains applicable even when there is limited training data available.

GPT3Mix \citep{yoo2021gpt3mix} is a technique that leverages a large-scale LM, such as GPT-3, to generate highly realistic synthetic text samples from a mixture of real samples and utilizes soft-labels predicted by LMs, effectively distilling knowledge from the large-scale LMs. \citet{sharma2023augmenting} used Llama2 to generate unpaired text for existing and new domains, which generates pseudo-labels for the generated utterances using a pre-trained RoBERTa model \citep{liu2019roberta}. The results for spoken semantic parsing improve further by using different generation strategies. \citet{samuel2023can} assessed how well GPT-4, an advanced LM, performed as a substitute for human annotators in low-resource reading comprehension tasks, evaluated its performance by comparing its effectiveness after fine-tuning, and examining the cost involved in annotation processes. 

\citet{latif2023can} used a 64-dimensional discrete audio representation generated by a vector-quantized variational autoencoder (VQ-VAE) as ChatGPT's audio context for data annotation, showcasing the potential of LLMs in speech emotion data annotation through experimentation. \citet{chowdhury2023generative} introduced a framework designed to improve the distributional robustness of reading comprehension models by employing generative models for dataset augmentation. GPT-3.5 is utilized to generate context based on questions, and T5 is employed to generate question-answering pairs. The research includes a comprehensive quantitative evaluation to assess the capability of the LLM to generate high-quality synthetic data for question-answering tasks. This study demonstrates the ability of the LLM to generate high-quality synthetic data for question-answering tasks. \citet{chen2023minprompt} introduced MINPROMPT, a data augmentation framework designed for open-domain question answering (QA). The framework incorporates an approximate graph algorithm and unsupervised question generation techniques. MINPROMPT aims to enhance the performance of QA models by leveraging these methods to generate additional training data. With the help of MINPROMPT, researchers can improve the robustness and accuracy of their QA systems.

\citet{kaddour2023text} attempted to fine-tune smaller models (student models) with training data generated or annotated by an LLM (Teacher model, e.g. GPT-NeoX \citep{black2022gpt}) to improve the downstream performance of much smaller models. To improve the performance of seq2seq automated audio captioning (AAC) models, \citet{wu2023improving} proposed a novel data augmentation method that uses ChatGPT to produce caption mix-ups (i.e., grammatical and compact combinations of two captions) , together with the corresponding audio mixtures, which increase not only the amount but also the complexity and diversity of training data.
CATfOOD \citep{sachdeva2023catfood} utilizes LLMs to enhance the training data of small language models (SLMs) through the generation of automatically altered inputs known as counterfactual (CF) instances. This approach aims to improve the out-of-domain (OOD) performance of SLMs in the extractive QA setup. 
\citet{ye2024llm} introduced LLM-DA, a data augmentation technique for few-shot Named Entity Recognition (NER) tasks. LLM-DA leverages LLMs to generate high-quality data by incorporating contextual rewriting, entity replacements, and noise injection, aiming to improve NER model performance in scenarios with limited training samples.

\subsubsection{Generated content-based approaches}
\citet{zheng2023augesc} used GPT-J \citep{gpt-j} to augment emotional support conversations (ESC) via a dialogue completion task. By prompting a fine-tuned LM with available dialogue posts from varied topics, the researchers generated full dialogues that are postprocessed using heuristics. Inspired by the recent success of LLMs, especially the development of ChatGPT, which demonstrates improved language comprehension abilities, \citet{dai2023chataug} proposed a text data augmentation approach based on ChatGPT (named AugGPT). AugGPT rephrases each sentence in the training samples into multiple conceptually similar but semantically different samples, ensuring both the correct labeling and sufficient diversity of the generated data. 

CoCa \citep{yu2022coca} represents a sophisticated representation learning approach that adeptly integrates natural language supervision. This is accomplished through pre-training on a diverse array of image-text data sourced from multiple origins. The approach effectively amalgamates contrastive and captioning losses within an encoder-decoder framework, showcasing its innovation in representation learning.
\citet{jo2022dagam} introduced a data augmentation technique called DAG, which utilizes T5-base \citep{raffel2020exploring} as the generation model to summarize a group of sentences from the original data to construct a longer sequence. This process generates augmented data with a representation distribution that is similar to the original data. 

\citet{oh2023data} undertook prompt-based data augmentation experiments utilizing ChatGPT, examining the diversity of data generated from three distinct prompts. The primary aim of this study is to refine the LM for an online call center's automatic speech recognition (ASR) system, specifically tailored for Hungarian, a language characterized by its complex morphology. The authors implemented a pre-training strategy, leveraging parliamentary text along with a GPT-2 transformer LM. This was followed by fine-tuning the model to align with the target domain and generating training text for a bidirectional neural language model (BNLM). The findings reveal that data augmentation via Transformer-based methods proves effective for the morphologically intricate Hungarian language, contingent upon a sufficiently extensive vocabulary and a robust BNLM. 
\citet{tarjan2020deep} pre-trained a GPT-2 Transformer LM on a general text corpus and fine-tuned it on the Hungarian conversational call center ASR task. Subsequently, this model is utilized to generate training text for a BNLM.

\citet{zhou2021flipda} proposed a novel data augmentation method FlipDA that jointly uses a generative model and a classifier to generate label-flipped data. Central to the idea of FlipDA is the discovery that generating label-flipped data is more crucial to performance than generating label-preserved data. Based on this observation, FlipDA first generates data using word substitution based on a pre-trained T5 and uses a classifier to select label-flipped data. 
\citet{guo2022genius} presented GENIUS, a text generation model that operates based on conditional input in the form of sketches. GENIUS is designed to fill in the missing contexts in a given sketch. Additionally, the study demonstrates that GENIUS can serve as a powerful and readily applicable tool for data augmentation in various NLP tasks. InPars \citep{bonifacio2022inpars}, a method for generating synthetic training data for information retrieval tasks, utilizes LLMs in a few-shot manner. It generates one question per document by employing GPT-3's Curie model, while using the "vanilla" and "guided by bad questions" (GBQ) prompt templates. 

\citet{khatri2022skillbot} illustrated that the capabilities of a substantial pre-trained transformer-based LM, such as GPT-2, can be effectively utilized to enrich limited datasets created by humans. This enhancement process is designed to preserve the original intent of the expanded utterances while also capturing various alternative expressions for the same intent. Consequently, this methodology leads to a notable improvement in the performance of chatbots driven by machine learning, enabling them to respond more accurately and diversely in conversational contexts. \citet{quteineh2020textual} introduced a new method of data augmentation that leverages the guided outputs of a language generation model such as GPT-2 to enhance the performance of text classifiers through an active learning process, which aims to generate synthetic data as unlabeled data that is required by active learning algorithms.

\citet{liu2022wanli} introduced a fresh method to create datasets by combining the efforts of human workers and AI. It began with an established dataset called MultiNLI, which is designed to deal with natural language inference. The proposed approach involves employing dataset cartography to automatically identify instances that exhibit complex reasoning patterns. These patterns serve as guidelines for instructing GPT-3 to generate new examples with similar characteristics. The generated examples are then filtered automatically and subsequently reviewed and labeled by human crowd workers. \citet{lu2023epa} proposed a novel method called easy prompt augmentation (EPA), which aims at improving the performance of LLMs via in-context learning with the paraphrasing ability of ChatGPT. This approach enables the automatic augmentation of task demonstrations by generating multiple paraphrased versions from various sources and targets.

FewGen \citep{meng2023tuning} utilizes few-shot samples to fine-tune a generator, generating data that enhances classification model generalization. It emphasizes label-discriminative information during tuning through a weighted maximum likelihood objective with automatically learned token weights. Inspired by the impressive text generation capabilities of modern pre-trained LMs, \citet{meng2022generating} presented supervision generation (SuperGen). In this approach, training data is generated by a unidirectional PLM, commonly referred to as the generator. Subsequently, a bidirectional PLM, known as the classifier, is fine-tuned on the generated texts to effectively carry out the associated task.
\citet{saakyan2023iclef} proposed a framework that utilizes model distillation from ChatGPT to enhance a formality style transfer dataset and provide explanations. Furthermore, a fresh method called in-context learning from expert feedback(ICLEF) is used to integrate scarce expert human feedback, prompting ChatGPT to evaluate its own outputs and refine them accordingly.

\citet{ko2023natural} introduced a score-based paraphrasing method that combines score evaluation with linear interpolation. This approach assigns Likert scale scores to sentences, reflecting their linguistic variation. The model then rewrites sentences using these scores, ensuring syntactic modification while preserving the original meaning, resulting in a seamless yet diverse linguistic transition. \citet{schlegel2023pulsar} utilized the LLMs by providing them with a medical note or a snippet and then prompted the LLMs to generate hypothetical conversations simulating interactions between a doctor and a patient. These generated conversations are used as the input to train the model in summarizing patient-doctor dialogues into clinical records.

\citet{gao2022self} introduced an innovative noise-robust re-weighting framework called SUNGEN. This framework aims to automatically generate top-notch data for zero-shot classification tasks. Specifically, the synthesized data produced by the PLM (GPT-2 XL) serves as a vessel of knowledge, which is utilized to train a task-specific model with significantly fewer parameters compared to the PLM. 
\citet{cai2023resolving} employed the LLaMA \citep{touvron2023llama} as a data generator to create high-quality scientific text data, aiming to address the challenge of imbalanced data and improve the performance of the model in classifying scientific texts. 
\citet{tang2023just} leveraged state-of-the-art LLMs, such as ChatGPT, GPT-4, Dolly-v2\citep{conover2023free}, and StableVicuna \citep{StableLM2024}, for the generation of synthetic data in the context of security patch detection. By prompting these LLMs, explanations for the patches are generated, with explicit instructions provided for binary-detection tasks. Dolly-v2-12b and StableVicuna13B are strategically employed to strike a balance between open-source and proprietary model contributions. The resultant dataset is organized in the format of $<$ patch, explanation, description, instruction $>$, encapsulating a comprehensive set of information for the purpose of security patch detection. 
\citet{li2023dail} introduced a method called data augmentation for in-context learning (DAIL). DAIL capitalizes on the idea that LLMs have a better understanding of the content they produce. Initially, it utilizes the LM to create paraphrases of the test sample and then uses majority voting to establish the ultimate outcome, considering individual predictions. \citet{ubani2023zeroshotdataaug} introduced an innovative technique employing zero-shot ChatGPT prompts for data augmentation in dealing with low-resource data scenarios. This method notably surpasses most baseline models on three text classification datasets: SST-2 \citep{socher2013recursive}, SNIPS \citep{coucke2018snips}, and TREC \citep{li2002learning}, showcasing its considerable potential in low-resource data settings.
\citet{wang2024large} suggested leveraging LLMs as tools for data augmentation to improve cold-start item recommendations by deducing user preferences from textual descriptions of new items, resulting in notable enhancements in recommendation accuracy for items lacking historical interaction data.

\subsection{Paired Data Augmentation}
\label{sec:PDA}
MixGen \citep{hao2023mixgen}, a data augmentation method for vision-language representation learning, generates image-text pairs with preserved semantic relationships through image interpolation and text concatenation. 
\citet{bakhtiarnia2023promptmix} proposed a method called PromptMix that extracts text descriptions from existing datasets, uses the extracted text as input to LDMs to generate images that are similar to those in existing datasets, annotates the generated images using high-performing heavy-weight networks, and mixes this fake dataset with real data to improve the training of light-weight deep neural networks.
To address the problem of reporting bias in visual language datasets, especially the potentially detrimental effect of object attribute associations on trained models, \citet{wu2023towards} proposed a bimodal enhancement method called BigAug. The BigAug utilizes object attribute decoupling to synthesize different visual language examples and create cross-modal hard negations. The integration of an LLM and a grounded object detector facilitates the extraction of target objects, where the LLM provides detailed attribute descriptions for each object. These descriptions, along with the corresponding hard negatives, are then used to generate images via the inpainted model. This explicit process introduces missing objects and attributes for learning, where the hard negatives guide the model to distinguish object attributes.

\section{Data Post Processing}
\label{sec-DPP}
Post-processing of augmented data is crucial for filtering out unsuitable instances. As shown in \tablename~\ref{table:dpp}, the post-processing techniques involve the application of various methodologies, including Top-$K$ selection, model-based approaches, score-based approaches, and cluster-based approaches. These post-processing techniques collectively contribute to refining the augmented dataset for optimal performance in subsequent tasks.
\begin{table*}[ht!]
\centering
\caption{Summary of data post processing including Top-K Selection, Model-based approaches, Score-based approaches and Cluster-based approaches (ordered by the name of the first author).}

\label{table:dpp}
    \begin{tabular}{|m{4cm}|c|c|c|c|c|c|c|c|c|c|}
    \hline
    Reference & Top-K Selection & Model-based approaches & Score-based approaches & Cluster-based approaches \\ \hline

\citet{bonifacio2022inpars} & \cmark & & & \\ \hline
\citet{chowdhury2023generative} & & & \cmark & \\ \hline
\citet{liu2022wanli} & & & \cmark & \\ \hline
\citet{meng2022generating} & \cmark & & & \\ \hline
\citet{sachdeva2023catfood} & & \cmark & & \\ \hline
\citet{sahu2022data} & & \cmark & & \\ \hline
\citet{samuel2023can} & & \cmark & & \\ \hline
\citet{shao2023diffuseexpand} & & & \cmark & \\ \hline
\citet{sharma2023augmenting} & & \cmark & & \\ \hline
\citet{vu2021strata} & \cmark & & & \\ \hline
\citet{wu2023image} & & & \cmark & \\ \hline
\citet{yu2023diffusion} & & & & \cmark \\ \hline
\citet{zhou2021flipda} & & \cmark & & \\ \hline
\citet{zheng2023augesc} & & & \cmark & \\ \hline

    \end{tabular}
\end{table*}

\subsection{Top-$K$ Selection}
Top-$K$ selection involves retaining the top-$K$ relevant and significant instances based on pre-defined criteria. In the final stage of constructing their training dataset, \citet{bonifacio2022inpars} selected the initial $K$ pairs based on log probability. Specifically, fine-tuning is limited to the top $K=10,000$ data pairs. Notably, fine-tuning across all 100,000 synthetic examples leads to a 4\% decrease in MRR@10 on the MS MARCO dataset, compared to the approach of filtering the top $K$ pairs. This finding underscores the importance of strategic data selection in model fine-tuning. 
To construct a training set, \citet{meng2022generating} adopted a strategy of generating a surplus of samples and then selecting the training data based on a scoring formula. For all tasks within GLUE \citep{wang2018glue}, except for the Corpus of Linguistic Acceptability (CoLA), the top $K$ samples from each class were chosen. Specifically for CoLA, the top $K$ samples were designated as linguistically acceptable training instances, while the bottom $K$ samples were categorized as linguistically unacceptable sequences. Moreover, to enhance fine-tuning stability and foster generalization, \citet{meng2022generating} implemented two regularization techniques: label smoothing and temporal ensembling. 
\citet{vu2021strata} developed a suite of methods encompassing overgeneration and filtering to augment both the quantity and quality of training data for synthetic NLI. In particular, a top-$K$ sampling technique, with $K$ set to 40, is employed to produce 100 output samples for each input, ensuring the removal of duplicates. Following this, a BERT model, meticulously fine-tuned on the MultiNLI (MNLI) dataset \citep{williams2017broad} in its original format, serves as an NLI classifier to sift through the synthetic training examples. The inclusion of a synthetic example is contingent upon the NLI classifier assigning the same label as that provided to the NLI data generator and demonstrating substantial confidence in its prediction.

\subsection{Model-based Approaches}
Model-based approaches leverage the knowledge and characteristics of the underlying models to refine and select augmented data. 
Despite the considerable generative capacity of LLMs, they tend to produce open-ended questions that cannot be resolved solely based on the input context. 

With this in mind, \citet{sachdeva2023catfood} used the Flan-UL2 model through context-generated QA pairs, aiming to determine the consistency between the context and the answer. The model makes a binary decision, labeling outputs as either "true" or "false." Consequently, instances, where the context lacks sufficient information, are eliminated.
To salvage questions mistakenly discarded due to contextual relevance filtering, \citet{sachdeva2023catfood} adopted a round-trip consistency approach \citep{alberti2019synthetic,fang2020accelerating}. This method employs existing QA models to answer questions generated by the LLM, ensuring the predicted answers align with the target answers generated by the LLM. In the noise filtering process, an ensemble of three LLMs, initialized with different random seeds during inference, is utilized. This technique maintains instances where at least two of the generated context-filtered questions (CFs) concur, successfully retaining between 90\% to 95\% of the data that would otherwise be discarded due to contextual relevance filtering as per the Duo-QAG method.

\citet{samuel2023can} used a round-trip filtration technique to improve the quality of synthetic question-answer pairs generated by GPT-4. It involves providing the question back to the model without the answer, allowing the model to attempt to answer the question again based on the context. If the model's newly generated answer matches the original synthetic answer, the QA pair is retained as it indicates a high-quality question with a consistent answer. If the answers do not match, the synthetic QA pair is discarded under the assumption that the question is flawed in some way. This helps to improve the quality of synthetic data, which in turn can improve the performance of downstream tasks. \citet{sharma2023augmenting} conducted a validation of the generated sequence logic parses, scrutinizing them for incorrect bracket placements and the occurrence of out-of-vocabulary (OOV) intents and slots. To rectify OOV intents, the model is re-prompted to replace them with the appropriate intents, ensuring that any intents beyond the first are substituted accordingly. In situations involving OOV slots, these are excluded from the sequence, while the corresponding slot words are retained.

To solve the problem that LLMs often generate utterances that belong to a closely-related intent rather than the desired one, \citet{sahu2022data} presented a prompting-based GPT classifier that acts as a filter to remove unfaithful examples and improves the quality of the training set. Specifically, the approach involves rejecting generated examples if they are identified by the GPT-3 classifier as not belonging to the seed intent. The fidelity of the generated data is greatly enhanced by applying this filtering method to both the HWU64 \citep{liu2019benchmarking} dataset and the Banking77 \citep{casanueva2020efficient} dataset. 
\citet{zhou2021flipda} introduced a data selection methodology that filters samples generated by a pre-trained T5 model using a classifier trained without data augmentation. This proposed method encompasses two steps. In the first step, generated candidate samples whose labels, as predicted by the classifier, differ from the original ones are eliminated. The second step involves categorizing the remaining candidates by their labels and selecting those with the highest probability within each group.
\citet{wu2023improving} used the FENSE disfluency detector \citep{zhou2022can} to remove the mixture of two audio titles with poor quality, which is produced by ChatGPT.

\subsection{Score-based Approaches} 
Score-based approaches assign scores to instances, allowing for the prioritization of those with higher relevance. DiffuseExpand \citep{shao2023diffuseexpand} utilizes a neural network to discern and retain high-quality Image-Mask pairs while filtering out those of lower quality. This process involves utilizing a well-trained neural network to evaluate the Dice loss associated with each synthesized image-mask pair. Samples exhibiting a Dice loss below a specified threshold $\eta$ are preserved, whereas those surpassing this threshold are excluded. This approach effectively eliminates pairs that are misaligned or inadequately synthesized.
\citet{wu2023image} proposed a high-quality synthetic image selection method. In the selection stage, the synthetic image is fed into a classifier, and then a global max-pooling (GMP) operation is applied to generate the image-level prediction score. Subsequently, classes with scores exceeding a specific threshold are considered as the actual labels for the generated image. If the actual label is a subset of the labels associated with the input image, then the generated sample is included in the synthetic dataset. 

\citet{chowdhury2023generative} screened QA pairs according to round-trip consistency \citep{alberti2019synthetic}, which is ensured by computing an auxiliary function greater than a set threshold.
\citet{zheng2023augesc} eliminated three types of undesirable situations: (1) failures in enhancement, encompassing the generation of non-dialogical content, incomplete dialogues, and cue word leakage; (2) harmful self-reinforcement, which targets the model's inclination to replicate patterns, particularly in instances of imbalanced corpus counts or consecutive speaker statements; and (3) distributional gaps concerning ESConv\citep{liu2021towards}. Additionally, criteria for the total number and length of utterances are established to minimize significant distributional gaps with ESConv and foster in-depth discussions with an ample number of dialogue rounds. These thresholds are determined based on both heuristics and ESConv statistics.
\citet{liu2022wanli} introduced an automatic filtering approach with the primary objective of selecting and retaining the most ambiguous examples from a generated dataset. Initially, this approach involves discarding failure examples produced by GPT-3 through a straightforward heuristic method. The process then introduces a novel metric, termed 'estimated max variability,' designed to assess the ambiguity of each remaining unlabeled example without necessitating additional training. This metric computes the maximum potential variability in the predicted labels for a given example. Following this, an equal number of examples exhibiting the highest max variability are retained from each intended label class, ensuring a balanced representation of ambiguity across the dataset.
The resulting filtered dataset, called $D_{filtered}$, will be half the size of the original generated dataset, $D_{gen}$. These filtered examples play a crucial role in training a more robust model that can effectively handle a wider array of inputs and reduce its tendency to overfit specific patterns in the training data.

\subsection{Cluster-based Approaches}
Cluster-based approaches aggregate similar instances, thereby aiding in identifying and eliminating redundant or less informative data. To evaluate the effectiveness of the proposed augmentation method for downstream segmentation tasks, \citet{yu2023diffusion} created four subsets from each training dataset. This process entails cropping images into $256\times256$ pixel patches, extracting features by using a pre-trained ResNet-50, classifying these patches into six groups through K-means clustering, and then selecting patches proximal to the cluster centers.

\section{Applications}
\label{sec-app}
Applying the aforementioned methods for data augmentation has proven to be highly effective in downstream tasks. These tasks encompass natural language processing, computer vision, and audio signal processing, demonstrating significant performance improvements obtained by large model-based data augmentations. \tablename~\ref{table:applications} provides a comprehensive summary and presentation of the existing methods.

\begin{table*}[ht!]
\centering
\caption{Summary of applying data augmentation to downstream tasks: NLP (Natural Language Processing, including TC (Text Classification), QA (Question Answering), MT (Machine Translation), NLI (Natural Language Inference), DS (Dialogue Summarising) and Others), CV (Computer Vision, including IC (Image Classification), SS (Semantic Segmentation) and OD (Object Detection)), ASP (Audio Signal Processing) (ordered by the name of the first author).}
\label{table:applications}
    \begin{tabular}{|m{4cm}|c|c|c|c|c|c|c|c|c|c|}
    \hline
    \multirow{2}{*}{Reference} & 
    \multicolumn{6}{c|}{NLP} & \multicolumn{3}{c|}{CV} & \multirow{2}{*}{ASP} \\ \cline{2-10}
    
    & {TC} & {QA} & {MT} & {NLI} & {DS} & {Others} & {IC} & {SS} & {OD} & \\ \hline

\citet{bonifacio2022inpars} & & & & & & \cmark & & & & \\ \hline
\citet{cai2023resolving} & \cmark & & & & & & & & & \\ \hline
\citet{chen2023minprompt} & & \cmark & & & & & & & & \\ \hline
\citet{chowdhury2023generative} & & \cmark & & & & & & & & \\ \hline
\citet{dai2023chataug} & \cmark & & & & & & & & & \\ \hline
\citet{du2023boosting} & & & & & & & & \cmark & & \\ \hline
\citet{dunlap2023diversify} & & & & & & & \cmark & & & \\ \hline
\citet{erfanian2024chameleon}& & & & & & & \cmark & & & \\ \hline
\citet{feng2024instagen} & & & & & & & & & \cmark & \\ \hline
\citet{gao2022self} & \cmark & & & & & & & & & \\ \hline
\citet{guo2022genius} & \cmark & & & & & & & & & \\ \hline
\citet{hemati2023cross}& & & & & & & \cmark & & & \\ \hline
\citet{jo2022dagam} & \cmark & & & & & & & & & \\ \hline
\citet{kim2023adversarial} & & & & & & & & & & \cmark \\ \hline
\citet{kumar2020data} & \cmark & & & & & & & & & \\ \hline
\citet{latif2023can} & & & & & & & & & & \cmark \\ \hline
\citet{li2023dail} & \cmark & & & & & & & & & \\ \hline
\citet{liu2022wanli} & & & & \cmark & & & & & & \\ \hline
\citet{lu2023epa} & & & \cmark & \cmark & \cmark & \cmark & & & & \\ \hline
\citet{lu2023wovogen} & & & & & & & & & \cmark & \\ \hline
\citet{meng2023tuning} & \cmark & & & & & & & & & \\ \hline
\citet{oh2023data} & & & \cmark & & & & & & & \\ \hline
\citet{sachdeva2023catfood} & & \cmark & & & & & & & & \\ \hline
\citet{samuel2023can} & & \cmark & & & & & & & & \\ \hline
\citet{samuel2023all} & & & & & & & \cmark & & & \\ \hline
\citet{saakyan2023iclef} & \cmark & & & & & & & & & \\ \hline
\citet{schlegel2023pulsar} & & & & & \cmark & & & & & \\ \hline
\citet{schnell2023generative} & & & & & & & & \cmark & & \\ \hline
\citet{sharma2023augmenting} & & & & & & & & & & \cmark \\ \hline
\citet{tarjan2020deep} & & & & & & & & & & \cmark \\ \hline
\citet{thakur2020augmented} & & & & & & \cmark & & & & \\ \hline
\citet{trabucco2023effective} & & & & & & & \cmark & & & \\ \hline
\citet{voetman2023big} & & & & & & & & & \cmark & \\ \hline
\citet{wang2024diffusion}& & & & & & & & & \cmark & \\ \hline
\citet{wang2024large}& & & & & & \cmark & & & & \\ \hline
\citet{wu2023image} & & & & & & & & \cmark & & \\ \hline
\citet{wu2023improving} & & & & & & & & & & \cmark \\ \hline
\citet{ye2024llm}& & & & & & \cmark & & & & \\ \hline
\citet{yin2023ttida} & & & & & & & \cmark & & & \\ \hline
\citet{yoo2021gpt3mix} & \cmark & & & & & & & & & \\ \hline
\citet{yu2023diffusion} & & & & & & & & \cmark & & \\ \hline
\citet{zang2023boosting} & & & & & & & \cmark & & & \\ \hline
\citet{zhang2023emit} & & & & & & & & \cmark & & \\ \hline
\citet{zheng2023augesc} & & & & & & \cmark & & & & \\ \hline

    \end{tabular}
\end{table*}

\subsection{Natural Language Processing}
Augmented text data plays a crucial role in enhancing the performance of NLP models across various tasks, including text classification (TC), question answering (QA), machine translation (ML), natural language inference (NLI), dialogue summarizing (DS), and others. By expanding and diversifying the dataset, text augmentation contributes to a deeper and more nuanced understanding of language variations and contexts.

\subsubsection{Text classification}
\citet{dai2023chataug} introduced a text data augmentation approach based on ChatGPT (AugGPT) to rephrase each sentence in the training samples into multiple conceptually similar but semantically different samples. This method enhances the performance of the BERT model across Amazon \citep{he2016ups,bao2019few,wang2022sentence} (+8.2\%), Symptoms \citep{medical_speech_transcription_and_intent} (+25.3\%), and PubMed20K \citep{dernoncourt-lee-2017-pubmed} (+4.3\%) datasets, which shows superior performance over other methods such as CounterFittedEmbedding \citep{mrkvsic2016counter,alzantot2018generating} and InsertWordByGoogleNewsEmbedding \citep{ma2019nlpaug}. 
\citet{jo2022dagam} proposed a data augmentation with generation (DAG) technique, in which T5-base is used as a generation model to augment the original data. Compared to the case without data augmentation, applying DAG improves the accuracy on AGNews \citep{zhang2015character} (+0.02\%), 20Newgroup \citep{lang1995newsweeder} (+0.3\%), TREC \citep{li2002learning} (+0.4\%) and R8 \citep{debole2005analysis} (0.26\%) datasets.
\citet{kumar2020data} explored three methods respectively using auto-regressive models (GPT-2), auto-encoder models (BERT), and seq2seq models (BART) for conditional data augmentation on text classification datasets. The BART-based method demonstrates the best performance across multiple datasets, improving accuracy on SST-2 \citep{socher2013recursive} (+5.04\%), SNIPS \citep{coucke2018snips} (+7.86\%), and TREC (+18.74\%).

GPT3Mix \citet{yoo2021gpt3mix} generates synthetic text using LLMs like GPT-3, achieving data augmentation and knowledge distillation by creating examples inspired by existing data and training smaller models with soft labels from the LLM. GPT3Mix significantly improves average classification accuracy for DistilBERT-base \citep{sanh2019distilbert} and BERT-base models, showing notable enhancements across various datasets: SST-2 (+18.7\%, +20.9\%), COLA \citep{warstadt2019neural} (+5.7\%, +7.9\%), TREC6 \citep{voorhees1999trec} (+11.3\%, +15.6\%), CR \citep{hu2004mining} (+11.2\%, +11\%), MPQA \citep{wiebe2005annotating} (+13.6\%, +12.9\%), and RT20 \citep{yoo2021gpt3mix} (+3.1\%, +6.2\%).  
Self-guided noise-free data generation (SUNGEN) framework \citep{gao2022self} is designed for automatic construction of high-quality synthetic datasets for zero-shot classification. It is compared with ZEROGEN \citep{ye2022zerogen}, a recent work emphasizing zero-shot learning through dataset generation. The method achieves an average accuracy improvement of 7.26\% using a 1-layer Bi-LSTM \citep{huang2015bidirectional} classifier and 2.8\% using DistilBERT-base across eight datasets, including IMDB \citep{maas2011learning}, SST-2, Rotten \citep{pang2005seeing}, Amazon, Yelp \citep{zhang2015character}, Subj \citep{pang2004sentimental}, AGNews and DBpedia \citep{zhang2015character}.
\citet{cai2023resolving} introduced a Hierarchical-disciplinary-structure Prompted Scientific Text Generator (HiPSTG), which utilizes LLMs (Llama V1) as data generators to enhance imbalanced hierarchical organized datasets for scientific text classification. This method has demonstrated remarkable performance improvement, as the effective utilization of 1000 synthesized samples results in a significant enhancement of the overall model performance, evident from increased scores across MicroF1, MacroF1, Recall, and Precision metrics.

\citet{guo2022genius} introduced a conditional text generation model using sketches as input (GENIUS), which is designed as a versatile and ready-to-use data augmentation tool for various NLP tasks. Based on GENIUS, GeniusAug and GeniusAug-f are proposed, improving the average accuracy by 2.29\% and 2.78\% respectively across six low-resource text classification datasets, including Huff \citep{misra2021sculpting}, BBC \citep{greene2006practical}, Yahoo \citep{zhang2015character}, 20NG \citep{lang1995newsweeder}, IMDB, and SST-2. In comparison to other data augmentation methods, including EDA, STA \citep{guo2022selective}, BackTrans \citep{silfverberg2017data}, MLM \citep{kumar2020data}, C-MLM \citep{kumar2020data}, and LAMBADA \citep{anaby2020not}, the proposed GeniusAug and GeniusAug-f approaches demonstrate superior performance.
FewGen \citep{meng2023tuning} leverages few-shot training samples to fine-tune a PLM for synthesizing novel training data aiming at enhancing the performance of few-shot learning. This method consistently outperforms previous few-shot methods without augmentation, such as LM-BFF \citep{gao2020making}, P-Tuning \citep{liu2023gpt}, and DART \citep{zhang2021differentiable}, with an average improvement exceeding 5\% across seven classification tasks from the GLUE benchmark \citep{wang2018glue}, including MNLI, QQP, QNLI, SST-2, CoLA, RTE, and MRPC. 
\citet{saakyan2023iclef} utilized the PAN 2022 dataset \citep{babakov2022large} to ascertain whether two texts originate from the same author. The $Alpaca_{IF\rightarrow F}$ model \citep{alpaca} is employed to extract explanations comprising informality attributes and evidence. Despite the exclusion of evidence in the simplistic approach, the resulting explanations achieve a classification performance of 0.60 AUC, indicating their prospective use as interpretable authorship features in future investigations.
For in-contextual learning (ICL), \citet{li2023dail} proposed a data augmentation technique called DAIL. DAIL augments test samples by generating multiple paraphrases and combines individual results through ensembling to derive the final prediction. DAIL outperforms standard ICL and alternative ensemble-based methods in comparative experiments, highlighting its superior efficacy. 

\subsubsection{Question answering}
\citet{chen2023minprompt} introduced a graph-based minimal prompt data augmentation framework (MINPROMPT) for few-shot question answering tasks, which combines graph-based algorithms and unsupervised question generation techniques to extract the most informative QA training samples from raw text. MINPROMPT effectively enhances the average F1 score across various QA datasets for two few-shot QA methods: Splinter \citep{ram2021few} (+8.7\%) and FewshotQA \citep{chada2021fewshotqa} (+1.3\%). In addition, as the number of few-shot QA training samples decreases, the improvement of MINPROMPT becomes more pronounced because MINPROMPT incorporates external prior knowledge that is not present in the actual training samples.
\citet{chowdhury2023generative} proposed generative data augmentation using LLMs, a novel framework to enhance QA model performance under natural distribution shifts. The approach involves supplementing real datasets with diverse data generated by generative models (GPT-3.5 and T5). Training the RoBERTa-base \citep{liu2019roberta} model on a combination of real data and data generated by GPT-3.5 and T5 leads to improvements in exact match (EM) and F1 scores across multiple datasets created by \citet{miller2020effect}, including SQUAD
 V1.1 (+1.7\%, +2.3\%), New Wikipedia (+1.2\%, +1.7\%), New York Times (+3.2\%, +2.5\%), Amazon (+0.7\%, +0.4\%), and Reddit (+1.6\%, +1.6\%). 
\citet{samuel2023can} proposed a GPT-4 based data augmentation approach to generate synthetic data for low-resource machine reading comprehension. Incorporating one-shot synthetic examples enhances the RoBERTa-base model's performance on both EM and F1 scores for the CovidQA \citep{moller2020covid} (+11.11\%, +13.69\%) and PolicyQA \citep{ahmad2020policyqa} (+2.1\%, +1.96\%) datasets. Similarly, the addition of two-shot synthetic examples results in improved EM and F1 scores for CovidQA (+11.47\%, +11.46\%) and PolicyQA (+0.89\%, +1.47\%) datasets.
\citet{sachdeva2023catfood} proposed CATfOOD, a novel method that leverages the capabilities of LLMs like Flan-UL2 \citep{tay2022unifying} and LLaMA to augment the training data of SLMs by generating counterfactual instances. This approach enhances the EM scores of the RoBERTa-base model with counterfactual instances across out-of-distribution (OOD) datasets, including SQUAD \citep{rajpurkar2016squad}, SQUAD Adversarial \citep{jia2017adversarial}, TriviaQA \citep{joshi2017triviaqa}, HotpotQA \citep{yang2018hotpotqa}, Natural Questions (NQ) \citep{kwiatkowski2019natural}, and BioASQ \citep{tsatsaronis2015overview}.

\subsubsection{Machine translation}
\citet{lu2023epa} introduced easy prompt augmentation (EPA), which is designed to enhance the performance of LLMs through the automatic augmentation of demonstrations. In machine translation tasks, compared to ChatGPT, EPA can significantly improve chrF++ \citep{popovic2015chrf} by up to 6 times in low-resource languages and up to 3 times in high-resource languages.
\citet{oh2023data} introduced a prompt-based data augmentation technique called storytelling. This method generates a story utilizing ChatGPT based on the original source language sentence. Each sentence of the generated story is then translated into the target language to create parallel data. This method enhances the BLEU score \citep{papineni2002bleu} of the mBART-50 \citep{tang2020multilingual} model by 0.68 when applied to the Korean-German pairs in the financial domain of the AI-hub \citep{aihub} multilingual colloquial parallel corpus.

\subsubsection{Natural language inference}
\citet{lu2023epa} introduced Easy Prompt Augmentation (EPA) to augment demonstration for LLMs in natural language inference (NLI). Experiments demonstrate that EPA is highly effective in enhancing performance in NLI tasks. The accuracy achieved by EPA surpasses that of ChatGPT on the SNLI \citep{bowman2015large} (+2.77\%) and MNLI \citep{williams2017broad} (+5.01\%) corpora with 3-shot in-context learning.
\citet{liu2022wanli} introduced a novel approach for dataset creation based on worker and AI collaboration. They utilized MultiNLI \citep{williams2017broad} as the original dataset to create a new dataset called Worker and AI NLI (WANLI). In comparison with MultiNLI, the RoBERTa-large \citep{liu2019roberta} model trained on the WANLI dataset exhibits a substantial increase in accuracy on the test set, including Diagnostics \citep{wang2018glue} (+4\%), HANS \citep{mccoy2019right} (+11\%), and ANLI \citep{nie2019adversarial} (+9\%).

\subsubsection{Dialogue summarising}
\citet{lu2023epa} presented Easy Prompt Augmentation (EPA), a demonstration augmentation approach for LLMs in dialogue summarization. Compared to ChatGPT, EPA achieves up to 0.79 improvement in ROUGE-L \citep{lin2004rouge} F1 scores on the SAMSum dataset \citep{gliwa2019samsum}.
PULSAR \citet{schlegel2023pulsar} is a framework that focuses on summarizing patient-doctor dialogues into clinical records, which utilizes domain-specific pre-training and data augmentation techniques to develop a specialized LM that can effectively convert patient dialogues into medical records. This framework uses an LLM to generate medical records from dialogues to augment the training data. Under conditions of data scarcity, the proposed data augmentation method significantly boosts performance across all metrics including ROUGE-1 (+1.77\%), ROUGE-2 (+1.81\%), ROUGE-L (+2.94\%), and ROUGE-LSum (+2.45\%) \citep{lin2004rouge} in task C (full-encounter dialogue2Note summarization) of MEDIQA-Sum 2023 \citep{MEDIQA-Sum2023}.

\subsubsection{Others}
\citet{zheng2023augesc} developed an innovative approach for dialogue augmentation and created an augmentation dataset named AUGESC for emotional support conversation (ESC). Through interactive human evaluations using two BlenderBot \citep{roller2020recipes} models, one fine-tuned on ESConv dataset \citep{liu2021towards} and the other further fine-tuned on AUGESC, it is demonstrated that AUGESC substantially enhances ESC performance in terms of Fluency, Identification, Comforting, and Suggestion. 
\citet{thakur2020augmented} introduced Augmented SBERT (AugSBERT), a method that utilizes a cross-encoder (BERT) to label additional input pairs, thereby expanding the training data for the bi-encoder (SBERT) in pairwise sentence scoring tasks. Across all in-domain datasets, AugSBERT outperforms the bi-encoder SBERT \citep{reimers2019sentence} by 1 to 6 points (Spearman’s rank correlation $\rho\times100$ and F1 score of the positive class). In domain adaptation tasks, AugSBERT exceeds SBERT up to 37 points (AUC(0.05) score).
\citet{bonifacio2022inpars} proposed InPars, a method leveraging the pre-trained models to generate synthetic data for information retrieval. It is observed that unsupervised models fine-tuned with InPars outperform models of the same size in OpenAI's Search API. For instance, T5 with InPars surpasses the larger Curie and Davinci models \citep{neelakantan2022text} by a considerable margin.
\citep{lu2023epa} introduced a demonstration augmentation method, easy prompt augmentation (EPA). In paraphrasing tasks, EPA demonstrates significant improvements in BLEU (+0.22) and ROUGE-L (+0.34\%) scores compared to ChatGPT on Quora Question Pairs (QQP) dataset \citep{gong2022diffuseq}.
\citet{wang2024large} leverages LLMs to generate augmented training signals for cold-start items in recommender systems. This method is applied to two well-established recommender system frameworks: NeuMF \citep{he2017neural} and SASRec \citep{kang2018self}. It is demonstrated that this approach significantly enhances the recall for both NeuMF and SASRec on the public Amazon review datasets.
\citet{ye2024llm} introduced LLM-DA, a data augmentation strategy that makes use of LLMs to bolster the few-shot Named Entity Recognition (NER) task. In scenarios with limited resources, our method consistently improves model performance compared to existing approaches. Specifically, on the CoNLL'03 dataset \citep{sang2003introduction}, the proposed method improves the micro-F1 scores by at least 30\% compared to any other existing method. Additionally, when applied to the FEW-NERD dataset \citep{ding2021few}, LLM-DA surpasses the micro-F1 scores of methods without data augmentation, achieving a performance increase of 5\%-15\% in both 5-shot and 10-shot setups.

\subsection{Computer Vision}
Augmented images improve performance in computer vision (CV) tasks like image classification (IC), semantic segmentation (SS), and object detection (OD). The diversified dataset enhances model accuracy and understanding of diverse visual contexts, enabling more robust and versatile applications.
\subsubsection{Image classification}
For few-shot classification, the application of SeedSelect \citep{samuel2023all} when fine-tuning the CLIP classifier outperforms multiple benchmark methods such as zero-shot CLIP \citep{radford2021learning}, CooP \citep{zhou2022learning}, Tip Adapter \citep{zhang2022tip}, CT \& SD (classifier tuning with images generated using SD) \citep{clark2024text}, and Textual Inversion \citep{gal2022image}. Notably, even with just one training image, SeedSelect consistently generates valuable, diverse, and superior augmentations compared to previous methods.
\citet{dunlap2023diversify} proposed a augmentation method, automated language-guided image augmentation(ALIA), for fine-grained classification tasks. Conducting domain generalization experiments on iWildCam \citep{koh2021wilds}, an animal Classification dataset, the results show that ALIA not only surpasses all baseline methods (‘+CutMix’ \citep{yun2019cutmix}, ‘+RandAug’ \citep{cubuk2020randaugment}, ‘+Txt2Img’, ‘+Real’ \citep{dunlap2023diversify}), exhibiting a remarkable 17\% performance boost compared to training solely on the original data, but it also outperforms the addition of an equivalent amount of real data. 
For the fine-grained classification task on CUB, a fine-grained bird classification dataset \citep{wah2011caltech}, ALIA demonstrates superior performance compared to all baselines except for when real data is added. These findings highlight that ALIA offers greater performance enhancements than existing data augmentation methods, even in scenarios without domain shifts. On waterbirds \citep{sagawa2019distributionally}, a constructed dataset with contextual bias,  when compared to other augmentation baselines, ALIA improves the class-balanced accuracy by 7\% and demonstrates a similar in-domain accuracy of other augmentation methods, and it outperforms all methods except for the real data baseline in terms of overall accuracy. 

\citet{trabucco2023effective} introduced a flexible data augmentation strategy, data augmentation by fusion (DA-Fusion), and investigated two approaches: model-centric leakage prevention and data-centric leakage prevention, for avoiding the leakage of Stable Diffusion's training data. The researchers conducted experiments on few-shot image classification tasks using three classification datasets including Leafy Spurge \citep{trabucco2023effective}, PascalVOC \citep{everingham2010pascal}, and COCO \citep{lin2014microsoft}. With Model-Centric Leakage Prevention, DA-Fusion outperforms the baseline which uses a standard data augmentation strategy including random rotations and flips by about 1.8 points on PascalVOC, 5 points on COCO, and 1 point on Leafy Spurge. Additionally, DA-Fusion outperforms Real Guidance \citep{he2022synthetic} in terms of overall performance. With Data-Centric Leakage Prevention, DA-Fusion outperforms the baseline by about 3 points on PascalVOC, 5.5 points on COCO, and 1.6 points on Leafy Spurge, as well as outperforms Real Guidance in all domains. \citet{yin2023ttida} proposed text-to-text-to-image data augmentation (TTIDA) for data augmentation. For in-domain classification, the experimental results on the CIFAR-100 dataset \citep{krizhevsky2009learning} show that TTIDA outperforms all methods that add synthetic images generated by different approaches (traditional image transformations, DCGAN \citep{radford2015unsupervised}, CycleGAN \citep{zhu2017unpaired}, StyleGAN \citep{karras2019style}) on each synthetic ratio (“+20\%”, “+50\%”, “+100\%”, “+200\%”, “+300\%”, “+400\%”, “+500\%”), and can achieve a maximum accuracy improvement of 3\%. For cross-domain classification, TTIDA performs better in all situations (different source domains or target domains) compared to not using synthetic data. 

DiffAug \citep{zang2023boosting} excels in classification, outperforming its competitors by a margin of 2.1\% to 11.1\% across eight evaluations on four datasets. Its data augmentation surpasses other methods, effectively addressing the deficiency of robust techniques in these approaches and thereby improving overall performance. The DiffAug-processed data exhibits reduced overlap between groups, leading to enhanced classification and clustering results by establishing clearer boundaries between different data categories. Lastly, the versatility of the DiffAug approach opens opportunities to enhance various unsupervised learning methods, particularly in the complex realm of effective data augmentation techniques for biological data.
\citet{koohpayegani2023genie} evaluated the impact of GeNIe on few-shot classification, long-tailed classification, and fine-grained classification. The classification experiments, covering both few-shot and long-tail distribution scenarios, highlight the effectiveness of GeNIe, particularly in categories with limited examples.
\citet{erfanian2024chameleon} presents Chameleon, a fairness-aware data augmentation system designed to mitigate the underrepresentation of minority groups. Leveraging advancements in foundation models, Chameleon efficiently employs these models for data repair, adding a minimal amount of high-quality synthetic data that adheres to the original data distribution. A race-predicting Convolutional Neural Network model was trained on the FERETDB dataset \citep{phillips1998feret}. By addressing the unfairness and lack of coverage in the dataset using Chameleon, the F1-Score on minority groups including Black (+0.36), Hispanic (+0.20), and Middle Eastern (+0.27) is improved. The findings indicate that the Chameleon system effectively reduces the model's unfairness in downstream tasks, with a slight decrease in model performance, which is deemed acceptable.
Prompt-guided cross domain generative augmentation (CDGA-PG) \citep{hemati2023cross} aims to diminish the discrepancies between domains by generating synthetic data, thereby enhancing model performance in domain generalization tasks. Experimental results demonstrate that CDGA-PG surpasses current leading domain generalization algorithms in the DomainBed benchmark \citep{gulrajani2020search} evaluation, achieving state-of-the-art performance on the PACS \citep{li2017deeper}, OfficeHome \citep{venkateswara2017deep}, and DomainNet \citep{peng2019moment} datasets.

\subsubsection{Semantic segmentation}
EMIT-Diff \citep{zhang2023emit} leverages ControlNet \citep{zhang2023adding} to generate synthetic medical images that retain vital characteristics while using edge information to guide the synthesis. The method demonstrates substantial improvements in medical image segmentation across diverse datasets, including Ultrasound breast \citep{al2020dataset} (+13.87\%), CT spleen \citep{antonelli2022medical} (+0.38\%), and MRI prostate \citep{antonelli2022medical} (+7.78\%). 
\citet{du2023boosting} enhanced ControlNet with lesion-specific visual and textual prompts for dermatoscopic image generation and showed superiority in improving skin lesion segmentation performance, surpassing Pix2PixHD \citep{wang2018high} by over 5\%. The effect of different amounts (e.g., 1K, 3K, 5K) of synthetic data on performance has also been explored, demonstrating that increasing the amount of data has the potential to improve the effectiveness of segmentation model training.
\citet{yu2023diffusion} presented a diffusion-based method for augmenting nuclei segmentation datasets by generating synthetic nuclei structures and histopathology images, which are then integrated with real data for segmentation model training. The effectiveness of the proposed augmentation method is evaluated by comparing the segmentation performance using both the original and augmented subsets for training two nuclei segmentation models, Hover-Net \citep{graham2019hover} and PFF-Net \citep{liu2021panoptic}. Remarkably, the experimental results demonstrate that augmenting as little as 10\% of the labeled real dataset with synthetic samples leads to segmentation performance on par with a fully-supervised baseline. 

In the training phase of Weakly-supervised semantic segmentation (WSSS), \citet{wu2023image} merged the synthetic dataset with the original training dataset to form the final training dataset. To evaluate the approach, it is incorporated into the existing state-of-the-art ViT-PCM \citep{rossetti2022max} as an upstream data augmentation technique, with consistency maintained in the downstream WSSS process. Subsequently, a comparison is made with segmentation results from various state-of-the-art methods. Notably, the proposed method demonstrates superior performance to the baseline ViT-PCM, even in the case when only 50\% of the training data is utilized.
\citet{schnell2023generative} examined two weakly-supervised semantic segmentation methods: simple regularized losses (RLoss) \citep{tang2018regularized}  and adaptive Gaussian mixture models (AGMM)  \citep{wu2023sparsely}. Both methods undergo joint training on the original and augmented training sets. The proposed framework effectively narrows the performance gap between scribble-supervised segmentation and fully-supervised segmentation. Furthermore, a notable improvement in segmentation performance on small datasets is demonstrated, surpassing the performance of even fully-supervised segmentation.

\subsubsection{Object detection}
\citet{voetman2023big} presented Genfusion, a framework that integrates recent developments to generate synthetic datasets. 
Synthetic training datasets are generated from a few real-world images using DreamBooth \citep{ruiz2023dreambooth} fine-tuning, these images are manually annotated. YOLOv5 \citep{Jocher_YOLOv5_by_Ultralytics_2020} and YOLOv8 \citep{Jocher_YOLO_by_Ultralytics_2023} are trained as object detectors. Evaluation is performed on the MinneApple benchmark \citep{hani2020minneapple}. Additionally, a preliminary investigation explores the potential for fine-tuning the diffusion model to automate annotation creation.
The performance of these models, evaluated on a real-world test set of 331 images, closely matches that of a baseline model trained on genuine data. Specifically, when applied to apple detection in orchards, the average precision deviation from the baseline ranges from 0.09 to 0.12. These findings underscore the viability of synthetic data generation as a pragmatic alternative for training deep models, reducing the imperative for extensive real-world data acquisition.

WoVoGen \citep{lu2023wovogen}, an innovative framework, addresses the difficulties linked to the creation of multi-camera driving scene videos, offering potential applications in augmenting datasets for autonomous driving. Through the utilization of a distinct 4D world volume, WoVoGen establishes consistency within the world and across sensors, overcoming challenges associated with diversity and adapting to changing lighting conditions. WoVoGen achieves superior experimental results on the nuScenes dataset \citep{caesar2020nuscenes}. Specifically, BEVDet \citep{huang2021bevdet} is utilized as the benchmark model, and the training and evaluation are conducted for 3D object detection tasks on both the original nuScenes dataset and the nuScenes dataset generated using WoVoGen. The results indicate that the data generated by WoVoGen significantly improved  3D object detection (mAP) from 34.9 to 36.2.
\citet{feng2024instagen} proposes InstaGen, a data augmentation pipeline designed to generate images with object bounding boxes for the purpose of training object detectors. In open vocabulary object detection, detectors trained on synthetic datasets generated by InstaGen exhibit outstanding performance on the COCO benchmark evaluation, improving the Average Precision (AP) by 4.5 compared to existing CLIP-based methods. In data-sparse object detection, the Faster R-CNN \citep{ren2015faster} trained with synthetic images, consistently enhances performance across different levels of real data availability. Especially with limited real data, the synthetic dataset becomes crucial, boosting the detector's AP by 5.2 when only 10\% of the real COCO dataset is used.
\citet{wang2024diffusion} presented a novel framework that leverages ControlNet to augment training data for object counting tasks. Employing the proposed method with three top-performing open-source models, including STEERER \citep{han2023steerer}, CUT \citep{qian2022segmentation}, and Generalized Loss \citep{wan2021generalized}, demonstrates that incorporating synthetic data into training significantly improves the performance of counting models across essential metrics such as Mean Absolute Error (MAE) and Mean Squared Error (MSE). Furthermore, the data-augmented models achieve state-of-the-art performance across all evaluated counting datasets, including ShanghaiTech \citep{zhang2016single}, UCF-QNRF \citep{idrees2018composition}, and NWPU-Crowd \citep{wang2020nwpu}.

\subsection{Audio signal processing}
Through synthesizing audio signals or corresponding text, data augmentation enhances the model's performance in audio signal processing tasks.
\citet{wu2023improving} proposed a data augmentation method employing ChatGPT to ‘mix-up’ pairs of captions in the Clotho dataset \citep{drossos2020clotho}, and generate more complex and diverse in-domain training data for automated audio captioning. Conducting an ablation experiment, the results show that the ChatGPT mix-up method improves the SPIDEr-FL score \citep{drossos2020clotho} by 1.9 or 2.4 points in the cases where other model components are different. Compared to other top methods such as \citet{xu2022sjtu} and \citet{ye2022automated}, the proposed approach is state-of-the-art regarding the SPIDEr-FL metric and demonstrates strong performance in other metrics such as METEOR \citep{denkowski2014meteor}, CIDEr \citep{vedantam2015cider}, SPICE \citep{anderson2016spice}, and SPIDEr (a combination metric using SPICE and CIDEr). 
For spoken semantic parsing (SSP), when unpaired text is lacking in current textual datasets, \citet{sharma2023augmenting} suggested prompting Llama2 to generate transcript-semantic parse data (unpaired text) for both existing and new domains. Employing the data generated by Llama2 with JAT (Just Add Text) and TTS (Text-to-Speech) can enhance the performance of SSP by a significant margin of 1.4\% and 2.6\% EM scores absolute for both existing and new domains.
\citet{latif2023can} evaluated the effectiveness of ChatGPT in annotating speech data for speech emotion recognition (SER). The results show that when using the data augmented by sampling with ChatGPT labels, the unweighted average recall (UAR) is obviously improved compared to using the actual IECMOAP \citep{busso2008iemocap} labels in both within-corpus and cross-corpus settings. Additionally, compared to previous studies, the UAR of the proposed method is about 2 to 3 points higher than the second place. 
\citet{tarjan2020deep} proposed an approach called subword-based neural text augmentation, in which GPT-2 is applied to generate augmentation data for an automatic speech recognition (ASR) model. This approach achieves a significant improvement on the word error rate (WER) of the online ASR system on Hungarian call center conversations, and the WER is reduced from 21.9\% to 19.6\%. In terms of WER and the ability to recognize out-of-vocabulary (OOV) words, the subword-based neural text augmentation method outperforms the original word-based data augmentation technique \citep{wang2019improving}. At the same time, it maintains a relatively small vocabulary size and low memory requirements for the system. 
\citet{kim2023adversarial} introduced a technique to produce high-fidelity respiratory sound samples, leveraging an audio diffusion model as a conditional neural vocoder. By employing a proposed adversarial fine-tuning approach, the generated samples are seamlessly integrated with authentic data to alleviate distribution inconsistencies between synthetic and real samples, which improves the performance of respiratory sound classification. On the ICBHI dataset \citep{rocha2018alpha}, the proposed method, adversarial Fine-tuning with synthetic samples, achieves a 2.24\% improvement in the ICBHI score \citep{rocha2018alpha} and up to 26.58\% improvement in accuracy for minority classes over the baseline, AST \citep{gong2021ast} fine-tuned without synthetic samples.

\section{Summary}
\label{sec-sum}
In this section, we present a consolidated overview of the key findings from our review in sections \ref{sec-appro}, \ref{sec-DPP}, and \ref{sec-app}.

Large model-based data augmentation remains a field filled with opportunities and challenges. This survey aims to comprehensively review the large model-based data augmentation approaches, the accompanying data post processing techniques, and applications in downstream tasks. By summarizing and analyzing current works, we identify successes and failures of current methods and discern new trends in large model-based data augmentation. Furthermore, we summarize the existing protocols and benchmarks used for evaluating large model-based data augmentation. Most importantly, these summaries can help in proposing new challenges and opportunities for future research.

\subsection{The Success and Failure of Large Model-Based Data Augmentation}
Large models in data augmentation exhibit a mix of successes and failures. The following summary, drawn from evaluations of current large model-based data augmentation methods, provides a concise overview of these outcomes.
\subsubsection{Achievements}
\begin{itemize}
    \item Large models exhibit a high level of competence in natural language understanding (NLU), as evidenced by their ability to excel in tasks such as text classification \citep{dai2023chataug}, question answering \citep{chen2023minprompt}, and natural language inference \citep{lu2023epa}, through the comprehension and interpretation of textual data with accuracy and precision.
    \item Large models display a remarkable aptitude for natural language generation (NLG), as evident from their capacity to excel in tasks like machine translation \citep{oh2023data} and dialogue summarizing \citep{schlegel2023pulsar}, where they adeptly generate coherent and contextually relevant textual outputs.
    \item Large models exhibit remarkable proficiency in generating images through prompts \citep{brooks2023instructpix2pix}\citep{sun2023imagebrush,nguyen2023visual}, extending their applicability to downstream tasks such as image classification \citep{samuel2023all}, semantic segmentation \citep{du2023boosting}, and object detection \citep{lu2023wovogen}. The generated data proves to be highly effective and impactful in enhancing performance in these subsequent tasks.
    \item Large models are proficient in handling complex audio data, demonstrating expertise in downstream applications such as automatic speech recognition (ASR) \citep{tarjan2020deep} , speech emotion recognition (SER) \citep{latif2023can} , spoken semantic parsing (SSP) \citep{sharma2023augmenting}, and automated audio captioning (AAC) \citep{wu2023improving}.
    \item Large models demonstrate the capacity to customize content generation while maintaining the given subject \citep{gal2022image,ruiz2023dreambooth,kumari2023multi}, showcasing their adaptability to a wide range of prompts.
    \item Large models facilitate the enhancement of data augmentation techniques, enabling the generation of more sophisticated and diverse content. This advancement significantly contributes to the improvement of performance in downstream tasks, underscoring the pivotal role of large models in designing data augmentation methods \citep{dai2023chataug,dunlap2023diversify,wu2023improving}.
\end{itemize}
\subsubsection{Limitations}
\begin{itemize}
    \item Data augmentations are susceptible to underlying models, such as potential misunderstanding of text prompts and complex relationships \citep{li2023blip}, sensitivity to typographic attacks \citep{avrahami2022blended}, difficulty in processing images containing text characters \citep{wei2023elite}, and challenges in handling multiple subject combinations \citep{kumari2023multi,ma2023unified}.
    \item Large models often encounter challenges in generating results that deviate significantly from the norm, particularly in scenarios where there is a substantial disparity between prompts and images \citep{tumanyan2023plug,sun2023imagebrush}, and in cases involving the synthesis of rare or highly fictional subjects \citep{ma2023unified}.
    \item Large models require precise prompts from users, but it is difficult to provide specific instructions for complex objectives \citep{hertz2022prompt}, and the effectiveness of the generated output is limited by the ambiguity inherent in natural language prompts \citep{gal2022stylegan}.
    \item Large models may exhibit biases in image generation \citep{brooks2023instructpix2pix,nguyen2023visual} and have strong priors towards certain subjects \citep{chen2023subject}.
    \item The text data generated by large models may potentially contain toxic or biased content, which cannot be fully assessed through either automatic or human evaluation \citep{zheng2023augesc}. Fine-tuning a large model within a specific domain can enhance the performance of data augmentation, but it requires a significant amount of resources \citep{kaddour2023text}.
    \item Currently, a universal text data augmentation method based on large models, akin to Rotation or Translation, that can be widely applied across diverse downstream tasks does not exist \citep{kumar2023advanced}. A general-domain LLM like ChatGPT may produce inaccurate augmentation results owing to its deficiency in domain-specific knowledge \citep{dai2023chataug}.
    
\end{itemize}

\subsection{Protocols and Benchmarks for Evaluation}
The evaluation methods for large model-based data augmentation can be divided into two categories: one involves assessing the efficacy of data augmentation methods based on changes in performance metrics of the corresponding downstream tasks; the other entails evaluating data augmentation methods by calculating quality metrics for the data generated through large model-based data augmentation. The second type, however, is relatively less explored and not widely applied.

Currently, the effectiveness of large model-based data augmentation is typically assessed using performance metrics of downstream tasks. For instance, in the fields of NLP, CV, and speech signal processing, the impact of data augmentation is measured by the improvement in model performance on corresponding downstream tasks. Classification accuracy is commonly used to evaluate the performance of text classification models and thereby assess the effectiveness of data augmentation \citep{dai2023chataug,jo2022dagam,yoo2021gpt3mix}. In Question Answering tasks, exact match (EM) and F1 are two prevalent metrics for evaluating data augmentation performance \citep{chowdhury2023generative,samuel2023can,sachdeva2023catfood}. For Machine Translation tasks, BLEU is used to assess precision, while CHRF++ comprehensively evaluates the quality of text generation. Both metrics can measure the enhancement of machine translation performance due to data augmentation \citep{oh2023data,lu2023epa}. In Dialogue Summarizing, common evaluation metrics include Rouge and its various optimized versions, which judge the quality of summaries by statistically analyzing the overlap of n-grams between the machine-generated candidate summaries and standard summaries \citep{schlegel2023pulsar,lu2023epa}. Additionally, in other NLP tasks like semantic textual similarity (STS), Spearman’s rank correlation is utilized to evaluate the impact of data augmentation on improving model performance \citep{thakur2020augmented}.

In Image Classification tasks, classification accuracy is commonly used to evaluate model performance \citep{dunlap2023diversify,trabucco2023effective,yin2023ttida}. In Object Detection tasks, model performance is typically assessed by calculating the average precision (AP) or mean average precision (mAP) across various categories \citep{voetman2023big,lu2023wovogen}. In Image Segmentation, dice or mean intersection over union (mIoU) are frequently used to measure the overlap between predicted and actual segmentations \citep{zhang2023emit,du2023boosting,yu2023diffusion,wu2023image}. Both of these metrics can be applied to evaluate the effectiveness of data augmentation methods.

In spoken semantic parsing (SSP) tasks, the EM score is commonly used as an evaluation metric \citep{sharma2023augmenting}. For automatic speech recognition (ASR) tasks, model performance is measured using the word error rate (WER) of online automatic speech recognition systems, as well as the ability to recognize out-of-vocabulary (OOV) words \citep{tarjan2020deep}. In automated audio captioning (AAC) tasks, the SPIDEr-FL is employed as a performance metric \citep{wu2023improving}.

Additionally, the quality of data generated by large model-based data augmentation methods can also be directly assessed using certain quality measuring metrics. 
In the field of text augmentation, various metrics are employed to evaluate the quality and relevance of generated data. For example, \citet{dai2023chataug} evaluate the augmented datasets generated by the proposed method, AugGPT, and other baseline methods using two metrics: cosine similarity and TransRate. The cosine similarity measures the similarity between the generated data and the test dataset, while the TransRate metric measures the learnability of the data. To assess the similarity between data generated by ChatGPT and the original data, \citet{ubani2023zeroshotdataaug} employed word overlap to determine the percentage of unique overlapping words between paired examples.

In the field of image augmentation, three prominent metrics for assessing the quality of generated images are the fréchet inception distance (FID) score \citep{heusel2017gans}, the CLIP score \citep{radford2021learning} and the DINO score \citep{caron2021emerging}.
The FID score measures the distance between the distribution of generated and real images in a feature space from the Inception network. A lower FID score indicates a higher similarity between real and generated images, suggesting better quality. This metric is extensively used in evaluating generative models, particularly in image synthesis \citep{zhang2023adding,kumari2023multi}. On the other hand, the CLIP score, leveraging the capabilities of the CLIP, assesses how well the generated images align with specific textual descriptions. This makes the CLIP score a valuable tool in assessing the performance of generative models, especially in tasks that require precise alignment between text and image content \citep{shi2023instantbooth,ge2023expressive,brooks2023instructpix2pix,tumanyan2023plug}.
Lastly, the DINO score assesses the preservation of structural and contextual elements in generated images, using the DINO-ViT self-similarity distance. Lower DINO score indicates better structural integrity, making this metric essential for maintaining the authenticity of image features during augmentation processes. The DINO score is extensively used due to its ability to evaluate the structural preservation in augmented images \citep{ruiz2023dreambooth,tumanyan2023plug,wei2023elite}.

\section{Grand Challenges}
\label{sec-challenge}
Although previous research on large model-based data augmentation has achieved numerous notable successes, this field remains in its nascent stages, with several critical challenges yet to be addressed. This section underscores these challenges and explores potential future research directions.

\subsection{Theoretical Understanding} 

The field of data augmentation currently lacks substantial theoretical research, often being perceived merely as a supplementary tool for enhancing model performance. Specific data augmentation approaches may increase accuracy, but these improvements generally hinge on the assumption that augmented data are label-preserving and do not alter the data distribution. However, these assumptions frequently do not hold in practical scenarios, potentially leading to noisy labels, shifts in data distribution, and subsequently, diminished performance or generalization. Moreover, large models are typically treated as black boxes. Gaining a deeper understanding of the characteristics that empower these models is crucial, especially in determining their reliability for processing sensitive data. A comprehensive and rigorous interpretation of these models is essential, not only to elucidate why certain augmentation techniques effectively improve model performance, but also to guide the selection or design of the most appropriate and effective methods for dataset expansion. Consequently, a critical future direction lies in developing theoretical support for data augmentation. This would involve establishing frameworks and principles to underpin the practical application of augmentation techniques, ensuring their effectiveness and suitability for diverse datasets and modeling challenges.

\subsection{The Number of Augmented Data}

An intriguing aspect of data augmentation is that the enhancement in training data quantity does not invariably correlate with a linear improvement in performance. Firstly, beyond a certain data threshold, further augmentation might actually impair performance. This phenomenon could be attributed to the fact that, although the quantity of data increases, its diversity may not. Secondly, there is a lack of theoretical guidance regarding the optimal size of training datasets. The decision on dataset size, suitable for specific tasks and models, is often based on empirical judgment and extensive experimentation. Researchers typically tailor dataset sizes to align with the specific models, training goals, and challenges in data collection. Thirdly, class imbalance can significantly distort data distribution, with the learning process frequently biased towards the majority class, leading to inadequate modeling of minority classes. Therefore, oversampling minority classes becomes crucial in data augmentation. However, oversampling essentially entails repeated sampling from the existing distribution, which might result in overfitting. Consequently, determining the appropriate amount of data generation for different classes is crucial to enhance model performance without compromising data diversity. This necessitates a strategic balance to ensure that the augmented data contributes effectively to the model's learning, without losing the variety essential for robust generalization.

\subsection{Multimodal Data Augmentation}

While several studies have explored paired data augmentation \citep{hao2023mixgen,bakhtiarnia2023promptmix,wu2023improving}, developing effective large model-based methods for multimodal data generation remains a challenge. Most existing works concentrate on augmenting a single modality, yet there lies significant potential in simultaneous multimodal data augmentation for various tasks, such as image captioning and speech recognition. Additionally, while paired data augmentation is predominantly inspired by large models and has the capability to enrich data patterns, introduce more diversity, and ensure fidelity in the generated data, the exploration of multimodal data augmentation techniques represents a significant and promising challenge for future research in data augmentation.

\subsection{Language and Vision Foundation Models}

The rise of artificial intelligence-generated content (AIGC), from Stable Diffusion to ChatGPT, has captured significant attention in both academic and industrial circles. The GPT family, particularly GPT-4 \citep{openai2023gpt4}, has demonstrated remarkable content generation capabilities and unexpected emergent abilities. However, to date, there is no equivalent 'vision foundation model' in computer vision that demonstrates comparable generalization across thousands of tasks. An intriguing approach is to use text generated by these models as prompts to create augmented images, leveraging the full diversity of text prompts to enhance the quality of the generated images. Furthermore, utilizing the knowledge emergence ability of these models as a bridge to develop similar emergent capabilities in vision foundation models presents an exciting challenge for future research.

\subsection{Automatic Data Augmentation}

Despite their effectiveness, current large model-based data augmentation approaches predominantly rely on manual design. The development of methods for automatically selecting suitable types of large model-based data augmentation remains relatively unexplored. While some approaches have proven effective in specific tasks or scenarios, their generalizability across different tasks is often limited. The exploration of techniques to automatically learn data augmentation strategies, or to search for an optimal augmentation policy tailored to specific tasks, could significantly improve the generalizability of augmented data.

\subsection{Robust and Consistent Data Augmentation}

Despite the promising outcomes of current large model-based data augmentation methods in practical applications, they are constrained by the potential lack of robustness and consistency in the generated data. For instance, certain data augmentation methods might alter the breed of a cat in an image, leading to erroneous classifications by the classifier, as demonstrated in \citep{trabucco2023effective}. In the realm of natural language processing, particularly for tasks like augmenting medical texts, LLM-based data augmentation can produce irrelevant sentences owing to ChatGPT's limited domain-specific knowledge. Consequently, it is crucial to tailor general-domain large models with domain-specific data when addressing particular tasks, ensuring the augmented data's relevance and accuracy.

\subsection{Trustworthy Data Augmentation}

In the process of data augmentation, ensuring the trustworthiness of the augmented data is paramount, especially when it is used to train large models. The presence of bias and toxicity in the training data can lead large models to generate content that significantly deviates from human preferences and standards. Consequently, there is a pressing need not only for generating trustworthy data but also for implementing reliable data augmentation approaches. This is particularly relevant for NLP applications, where a major concern is how to rephrase sentences to convey high-level information without incorporating offensive content. Currently, there is a notable gap in research addressing this issue. Future work should focus on developing both trustworthy data augmentation techniques and robust evaluation frameworks for augmented data, ensuring that they adhere to ethical standards and reflect the desired level of quality and reliability.

\subsection{The Instruction Following Ability of Large Models}

The burgeoning success of large models has catalyzed a burgeoning interest in the study of prompt engineering. To generate high-quality augmented data, carefully crated prompts are critical. However, a notable challenge arises as large models often struggle with adhering to instructions, frequently generating outputs that deviate from the specified directives. Additionally, large models exhibit limitations in comprehending instructions characterized by complex structures or ambiguous goals. The challenge of accurately assessing the capacity of large models to follow instructions remains an unresolved issue, with a paucity of benchmarks dedicated to this purpose. Consequently, the development of robust evaluation protocols to measure the instruction-following capabilities of large models is of paramount importance, promising to advance the design of large model-based data augmentation methods.

\subsection{The Evaluation of Augmented Data}

The quantity of data generated by augmentation approaches is critically important. However, currently, there are no standardized evaluation metrics specifically for augmented data, making its quality assessment a major challenge. Presently, the quality of augmented data is typically assessed based on task-specific performances, such as evaluating data augmentation methods by their impact on tasks like text classification, measured by accuracy, or semantic segmentation, gauged by IOU scores. Yet, these do not provide direct metrics for the augmented data itself. Ideally, evaluation metrics should measure both the diversity of individual data points and the overall consistency of the dataset, independent of the specific task at hand. Moreover, it appears impractical to expect one or a few general datasets to capture the nuances of all data augmentation methods, particularly those tailored to specific tasks. Nonetheless, a small benchmark capable of evaluating various data augmentation approaches would be highly beneficial. Such a benchmark should assess different aspects of data augmentation methods, including diversity and faithfulness. With the increasing reliance on large models, it may also be prudent to use data that are not part of the training sets for these models. Incorporating testing data from these models might lead to inaccurate conclusions. Consequently, the development of such metrics and datasets is vital for the progression of data augmentation techniques, providing a clearer understanding of their effectiveness and applicability across different contexts.

\subsection{Beyond Augmentation: Training Large Models Using Augmented Data}

Data augmentation, while pivotal, serves merely as a starting point rather than the ultimate goal in the realm of machine learning. Training data, algorithmic innovation, and computational power are the triad underpinning the performance of large models. With these models rapidly advancing in capability, the scarcity of high-quality data is emerging as a primary bottleneck in scaling large models. This scenario underscores the importance of leveraging data generated by large models for training purposes. To optimize the use of such data, it is imperative to develop effective metrics that assess the diversity and faithfulness of the augmented data, thereby preventing model overfitting. A comprehensive data augmentation system should encompass not just metrics evaluating specific attributes of augmented data, such as diversity, but also a robust theoretical framework that elucidates the usefulness of this data. In conclusion, data augmentation holds the potential to address the challenge of data scarcity in training large models. There is significant scope for future advancements in this area, with the aim of enhancing the efficacy and understanding of data augmentation techniques.

\section{Conclusion}
\label{sec-conclusion}

Data augmentation holds profound significance, emerging as a crucial component in the advancement of artificial intelligence models, particularly in the context of large models. This survey offers an exhaustive examination of data augmentation methods driven by large models. We dissect and review these studies across three dimensions: approach, data post-processing, and application. For each dimension, we construct a detailed taxonomy to interlink existing research, summarizing key techniques and clarifying their strengths and limitations. Beyond reviewing past work, this survey also identifies several challenges within the field, poised to steer prospective future research directions.

\section*{Disclarimer}

The goal of this paper is mainly to summarize and discuss existing data augmentation methods driven by large models. Results and conclusions in each paper are original contributions of their corresponding authors.
This paper may discuss some side effects of large model-based data augmentation methods and the only intention is to foster a better understanding.

Due to the evolution of large models, it is very likely that they become stronger and some of the limitations described in this paper are mitigated (and new limitations may arise).
We encourage interested readers to take this survey as a reference for future research and conduct real experiments in current systems when performing data augmentation.

Finally, the field of data augmentation is continuously developing, thus we may miss some new papers. 
We welcome all constructive feedback and suggestions.


%

\ifCLASSOPTIONcaptionsoff
  \newpage
\fi

\bibliographystyle{apalike}
\bibliography{refs} 

\end{document}